\documentclass[%
	3p
]{article}

\usepackage{xcolor}
\usepackage{theStyle}
\usepackage{tikz}
\usetikzlibrary{shapes, arrows.meta, positioning}
\usepackage{overpic}

\usepackage{authblk}

\begin{document}

\title{
Inverting the Fundamental Diagram and Forecasting Boundary Conditions:
How Machine Learning Can Improve Macroscopic Models for Traffic Flow
}

\author[a]{Maya Briani}
\author[a]{Emiliano Cristiani}
\author[a,b]{Elia Onofri}

\affil[a]{\footnotesize Istituto per le Applicazioni del Calcolo, Consiglio Nazionale delle Ricerche, Rome, Italy}
\affil[b]{\footnotesize Dipartimento di Matematica e Fisica, Università degli Studi Roma Tre, Rome, Italy}

\date{\today}

\maketitle

\begin{abstract}

In this paper, we aim at developing new methods to join machine learning techniques and macroscopic differential models for vehicular traffic estimation and forecast.
It is well known that data-driven and model-driven approaches have (sometimes complementary) advantages and drawbacks. 
We consider here a dataset with flux and velocity data of vehicles moving on a highway, collected by fixed sensors and classified by lane and by class of vehicle.
By means of a machine learning model based on an LSTM recursive neural network, we extrapolate two important pieces of information: 1) if congestion is appearing under the sensor, and 2) the total amount of vehicles which is going to pass under the sensor in the next future (30 min).
These pieces of information are then used to improve the accuracy of an LWR-based first-order multi-class model describing the dynamics of traffic flow between sensors. The first piece of information is used to invert the (concave) fundamental diagram, thus recovering the density of vehicles from the flux data, and then inject directly the density datum in the model. This allows one to better approximate the dynamics between sensors, especially if an accident happens in a not monitored stretch of the road. 
The second piece of information is used instead as boundary conditions for the equations underlying the traffic model, to better reconstruct the total amount of vehicles on the road at any future time. 
Some examples motivated by real scenarios will be discussed. 
Real data are provided by the Italian motorway company Autovie Venete S.p.A.

\noindent\textbf{MSC-2020.} 76A30; 68T07. 
\end{abstract}

\section{Introduction}\label{sec:introduction}
Traffic state estimation (TSE) and traffic forecast have a long and solid tradition which dates back to the 1950s. A very broad division of the research lines on vehicular traffic flow modeling is summarized in the following diagram:
\begin{center}\scalebox{.6}{

\begin{tikzpicture}[
        theNode/.style={
            draw,
	        minimum width=2cm,
	        minimum height=0.5cm,
	        align=center
        },
        font=\small,
        thick
    ]
    
	\node[theNode] (block1) {traffic flow models};
	\node[theNode, below  left=of block1 ] (block9)  {model-driven};
	\node[theNode, below  left=of block9 ] (block2)  {differential};
	\node[theNode, below right=of block9 ] (block10) {nondifferential};
    \node[theNode, below right=of block1 ] (block3)  {data-driven};
	\node[theNode, below  left=of block2 ] (block4)  {micro};
	\node[theNode, below      =of block2 ] (block5)  {meso};
	\node[theNode, below right=of block2 ] (block6)  {macro};
	\node[theNode,       right=of block10] (block7)  {parametric};
	\node[theNode, below right=of block3 ] (block8)  {nonparametric};

	\draw[-latex] (block1) -- (block9);
	\draw[-latex] (block9) -- (block2);
	\draw[-latex] (block1) -- (block3);
	\draw[-latex] (block2) -- (block4);
	\draw[-latex] (block2) -- (block5);
	\draw[-latex] (block2) -- (block6);
	\draw[-latex] (block3) -- (block7);
	\draw[-latex] (block3) -- (block8);
	\draw[-latex] (block9) -- (block10);

\end{tikzpicture}         

}\end{center}
Model-driven and data-driven approaches have their own advantages and drawbacks, which were well described in, e.g., \cite{thodi2022IEEE-TITS, xiangxue2019AJSE}: Model-driven approaches allow to inject in the simulator the human knowledge of the system, at least if it can be reasonably translated into equations. 
Differences between agents can be (stochastically) taken into account as well, including drivers' psychological aspects. Differential macroscopic models, in particular, can unveil the power of methodologies based on partial differential equations (PDEs), for example giving the right tools to compute the \emph{Wardrop equilibrium} of a traffic system on a road network \cite{wardrop1952PICE, carlier2012JMS, cristiani2015NHM}.
On the other hand, this approach tends to be an over-simplification of traffic physics since the model is never able to catch \emph{all} the features of cars and drivers. 
Most models are difficult to work with noisy and fluctuated data collected by traffic sensors
and the calibration of the numerous parameters is quite challenging. 
In addition, the numerical scheme used for the discretization of the equations (Godunov, Lax-Friedrichs, etc.), actually needed to solve them, introducing a further, often not negligible, approximation error.
Finally, models require additional inputs which are not available in real scenarios, such as, e.g., boundary conditions at any future time. 

Data-driven approaches, instead, are more suitable to deal with the nonlinearities which characterize traffic flow and, for this reason, can be more accurate than model-driven approaches, but they are agnostic to the physics of traffic flow and could lead to infeasible estimation results. 
These methods are also not often interpretable and lack robustness. 
More importantly, the generalizability of the models is often weak and they have a high dependence on the training data samples. 
If the quality of training data is poor (missing/overestimated/underestimated data), their predictive accuracy will be severely weakened.
Recently, many machine learning (ML) approaches were proposed: they often rely on relatively simple structures (compared to PDEs intricate systems) making them more lightweight from a computational point of view, hence being more suitable for real-time applications.
However, dependency on large sets of historical data means that training can be computationally very expensive and can easily lead to data overfitting.

Recent research is growing in interest toward \emph{hybrid approaches} which try to get the best from each of the two approaches;
however, they are rarer in the literature and each paper uses only single aspects from the two methodologies to obtain very different results.
This paper tries to advance in this research field by proposing a computational method where ML techniques extrapolate from data the information needed by differential macroscopic models for traffic flow.

\subsection*{Relevant literature}

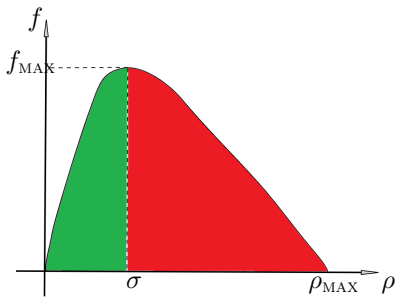
\begin{figure}[h!]
 \centering
 \begin{overpic}[width=.4\linewidth]{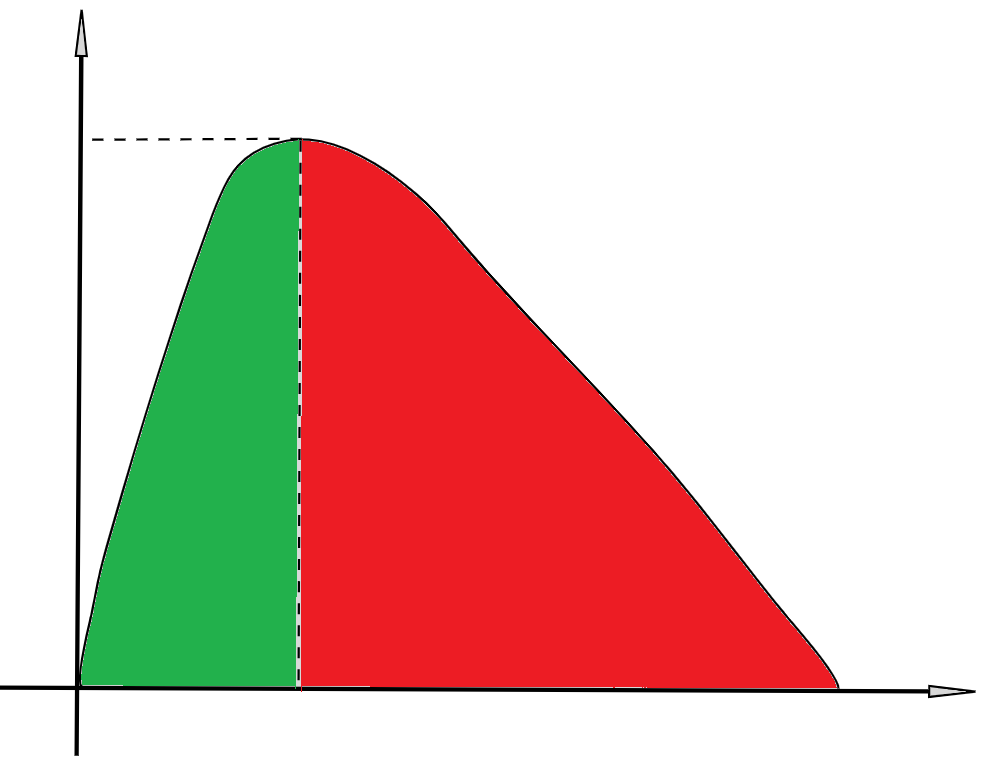} 
 \put(30,3){$\sigma$} \put(80,3){$\rho_{\textsc{max}}$} 
 \put(100,3){$\rho$} 
 \put(3,75){$f$} \put(-3,63){$f_{\textsc{max}}$} 
 \end{overpic} 
\caption{Fundamental diagram $f=f(\rho)$. The green part corresponds to the free phase while the red part corresponds to the congested phase. Density $\sigma$ corresponds to the maximal flux $f_{\textsc{max}}$ (road capacity). Flux is null when the road is empty ($\rho=0$) and when the road is fully congested  ($\rho=\rho_{\textsc{max}}$) and vehicles are stopped.}
 \label{fig:FDgeneric}
\end{figure}

\paragraph{Fundamental diagram}
First of all, we have to mention the \emph{fundamental diagram}, which is one of the basic ingredients of all model-driven approaches, especially at the macroscopic scale. 
It defines the relationship between the flux and the density
of vehicles \cite{kesselsbook, ferrarabook}, see Figure~\ref{fig:FDgeneric}. 
It is plain that the flux of vehicles is null in either the case of an empty road (null density) or in the case of a fully congested road (maximal density, stopped bumper-to-bumper vehicles). For intermediate density levels, real data show more complex dynamics, especially in correspondence to the maximal capacity of the road. 
Indeed, drivers act differently in response to the same traffic conditions and, in addition, accelerations and decelerations are far from being instantaneous. 
As a consequence, traffic shows some instabilities \cite{briani2021AXIOMS, ni2018AMM, wang2013JAT}.
In first-order traffic models, the fundamental diagram can be defined by means of a single function while second-order traffic models allow the fundamental diagram to be multivalued, in the sense that a single value of the density can be associated with
many values of the flux, exactly as it happens in reality.
Recently, an interesting ML-based method to estimate the fundamental diagram was proposed in \cite{shi2021IEEE-ITS}.

\paragraph{Model-driven} Mathematical models for traffic flow first appeared in 1955-6 with the seminal papers \cite{lighthill1955PRSLA, richards1956OR}, which introduced the well-known LWR model.
It is a first-order (velocity-based) model constituted by a hyperbolic PDE where the observed quantity is the vehicle density $\rho$.
The model stems from the reasonable analogy between vehicular dynamics and fluid dynamics.
Following the same line, the model was successively extended to the second-order to include inertia (acceleration) effects in \cite{P,W} and \cite{AR,Z}, giving birth to the PW and ARZ models, respectively. Independently, in the engineering literature, the CTM model \cite{CTM} and the METANET model \cite{METANET} were introduced.
These two last models are equivalent to the discretized versions of the LWR and PW model, respectively \cite{wang2022TRC}.
The literature about mathematical models is huge, we refer the reader to the surveys \cite{wang2022TRC, hoogendoorn2014EUROJTL, helbing2001RMP, bellomo2011SR, seo2017ARC} for differential models and their calibration, and \cite{maerivoet2005PR} for nondifferential models (cellular automata). 

Some effort was also devoted to generalizing mathematical models to road networks. This is not a trivial task due to the junction conditions which must be added to assure the uniqueness of the solution of the resulting system of PDEs. We refer the reader to \cite{briani2014NHM, bretti2014DCDS-S}, and the book \cite{piccolibook} for basic concepts.

Another generalization of our interest is that of multi-class models. In this case more than one class of vehicles (e.g., cars and trucks) share the same road. Each class has a specific dynamic and the classes interact with each other in a nontrivial manner.
We refer to \cite{fan2015SIAP, kessels2016TRR, qian2017TRB}, and to the recent books \cite{ferrarabook, kesselsbook} for an overview of the most used multi-class models.

\paragraph{Data-driven} Traffic data can be collected by means of several methods and technologies.
Commonly one can have Eulerian data, provided by fixed sensors placed along the road (which count passing vehicles), and Lagrangian data, provided by probe vehicles equipped, e.g., with a GPS system. 
We refer to \cite{chen2015IEEE-TITS, chow2015JTE, xing2022PhA} for an overview on traffic data.
Typical objects under analysis are, hence, flux $f$ and velocity $v$, as opposed to the more abstract concept of density $\rho$ that characterizes the model-driven approaches.

Early works in data-driven mainly rely on statistical approaches based on historical data.
More recently, the broad research carried out in ML furnished new lymph to data-driven approaches, with a large corpus of research exploiting from the simplest to the more exotic technique to study TSE problems.

In this paper, we are mostly interested in Artificial Neural Network (ANN) methods for traffic understanding, estimation, and prediction, see \cite{thodi2022IEEE-TITS, du2021SCS, modi2022ESWA, yan2022FGCS, zhang2020TRR, zheng2022ESWA}.
ANNs mainly divide into two families: feed-forward models, like the well-known Single- or Multi-Layer Perceptron (S/MLP) and feed-back models, like Recursive Neural Networks (RNNs).
We refer the reader to the recent surveys \cite{lee2021IEEE, wang2022IEEE-TKDA, fang2022PhA, manibardo2021IEEE-TITS, tedjopurnomo2020IEEE-TWDA} on this topic and, more specifically, we focus on the use of Long Short-Term Memory RNN (LSTM-RNN, or LSTM in short) for traffic data forecast.
LSTM is a powerful tool that proved to be capable of capturing long-range temporal feature dependencies and reducing gradient explosion/vanishing.
Among the recent literature, LSTM were often used (both vanilla or as a building block of more complex structures) to perform analysis and prediction on $f$ and $v$:
for what concerns velocity, it is the case, e.g., of \cite{gu2019TRC} that mounts a fusion deep learning approach to predict lane-level traffic speed at two minutes, \cite{hsueh2021IJITSR} that considers the correlation between car speed and car type for a prediction model (LSTM + 4 layer MLP) of the highway speed at 5 minutes, and \cite{xiangxue2019AJSE} that combines LSTM with a careful data preprocessing aided with wavelets analysis to perform speed prediction at 15 minutes.
For what concerns flux, it is the case, e.g., of \cite{zhao2017IET} where a temporal-spatial correlation is integrated into a 2D LSTM network to predict traffic flow at 15 minutes, \cite{wang2021PhA} which integrates weather data with an attention model to perform short-term prediction of the traffic volume, and \cite{fang2022PhA} which describes a novel methodology of an LSTM-based attention model to predict the upcoming flux based on 120 minutes of data (aggregated 10 minutes by 10 minutes).

\paragraph{Hybrid methods} To overcome the shortcomings of both model- and data-driven approaches while exploiting their potentialities, recent studies introduced coupled methods where physics and data play together.
The way the coupling is performed in the literature is very different since a common line is yet to be established:
the physical model can be 
(i) injected in the training process of the ANN, obtaining the so-called Physics-Informed Neural Networks (PINNs), 
(ii) used in parallel with the ANN, as it is done, e.g., in \cite{shi2021IEEE-ITS}, where the TSE, the model parameter identification, and estimation of the fundamental diagram are performed simultaneously,  
or (iii) used \emph{after} the ANN, like in the present work and in \cite{herty2022pp}, where data are used to provide consistent boundary conditions at junctions for macroscopic traffic flow models.

The majority of the recent works fall back into the PINNs category, where physics is usually plugged into the model by building a custom cost function, in particular, trying to exploit the powerfulness of deep learning models (PIDL); it is the case, e.g., of \cite{shi2021proc} where authors focus on highway TSE with observed data from loop detectors and probe vehicles, by building a coupled model with a Physic-Uninformed Neural Network (PUNN) and a PINN with custom loss function based on physical discrepancy measures.
Paper \cite{huang2022ITS}, analogously, builds a custom loss function relying on CTM and LWR with different fundamental diagrams (Greenshields', Daganzo's, and inverse-lambda) to tackle the problem of data sparsity and sensor noise.
Another example of custom cost function based on multiple physical aspects is given in \cite{barreau2021IEEEproc}, where authors perform TSE from probe vehicles data in an urban environment by building a 6-component loss function that is used both to reconstruct the road state and a smoothed version of the probe trajectories.
Also worth mentioning is the paper \cite{mo2021TRC}, where authors introduce PIDL car-following model architectures encoded with different popular physics-based models 
to predict the evolution of each vehicle's velocity.
Finally, it is also interesting the approach based on physics regularized Gaussian process (PRGP), like the one proposed in \cite{yuan2021TRB} (later extended in \cite{yuan2022pp}) where a stochastic PRGP is developed and a Bayesian inference algorithm is used to estimate the mean and kernel of the PRGP itself.

\subsection*{Paper contribution}
In this paper, we deal with traffic data coming from a series of fixed sensors placed along a highway. 
Sensors are about 1-to-20 km away from each other and are able to count vehicles passing under them, estimate vehicles' velocity, and classify vehicles in terms of their length (dividing them, e.g., between light and heavy vehicles).
Sensor data are aggregated every minute, sent to a central processing unit, and stored in a database. 
The velocity datum corresponds to the mean of the velocities of vehicles of each type observed in the time interval of 1 minute.

The main goal is to estimate the traffic conditions \emph{all along the road} at \emph{current} and \emph{future} time -- in terms of macroscopic quantities like flux $f$, density $\rho$, and velocity $v$ -- by means of a first-order multi-class LWR-like macroscopic differential model which describes the joint dynamics of light and heavy traffic, already introduced in \cite{briani2021AXIOMS}. 
More in detail, we consider 2 steps:

\medskip

\textbf{Nowcast.} This is the traffic estimation at current time $t_0$ (now), at every point of the road. To do that, we split the road into several consecutive segments, each of which starts and ends with a sensor. Then, we run the model setting the initial time $t=t_0-\Delta \tpast$ and the final time $t=t_0$, where $\Delta \tpast$ is a parameter. 
The model runs in each segment independently and gives the evolution of the density there. At time $t_0-\Delta \tpast$ we assume that the road is empty, then the road starts filling thanks to the sensor data which act as inflow and outflow boundary conditions. 
If $\Delta \tpast$ is sufficiently large, the road fills completely and a reliable density estimate is computed along each segment.

The problem arises how to employ sensor data to enforce boundary conditions: mathematical models typically require \emph{density data} as Dirichlet boundary conditions, but in our case sensors do not provide this information. 
Alternatively, one can inject the \emph{flux data} directly into the numerical scheme chosen for the discretization of the modeling equations. Unfortunately, this solution is not always feasible because sensor data are not guaranteed to be compatible with the solution of the numerical model;
moreover, the solution is not always the ``correct'' one, especially in the case of congestion events appearing between sensors, see Section \ref{sec:models} for details. 
This is the reason why we explore a third approach: we train an ANN based on an LSTM to detect congestion formation at sensors in real time. This is an interesting and complex problem \textit{per se}, which gives, as a by-product, a tool for \emph{inverting the concave fundamental diagram without ambiguity}, being able to distinguish the free phase from the congested phase, see Figure~\ref{fig:FDgeneric}. The tool is then used to transform the flux datum into a density value, and inject it into the model, thus solving the incompatibility issues mentioned above and also adding new physical information in the model, leading, in most of the cases, to a more accurate solution. 

\medskip

\textbf{Forecast.}  This is the traffic estimation at any future time $t_0<t<t_0+\Delta \tfut$, where $\Delta \tfut$ is the duration of the simulation (30 min, in our case). In this case, sensors data are not yet available, hence we forget the sensors and we consider the whole highway as a unique long segment. We employ the same mathematical model considered before, using the nowcast traffic estimation as initial conditions for the density. 

Since the model needs the boundary conditions for any time $t_0<t<t_0+\Delta \tfut$, i.e.\ it needs to have an estimation of the number of vehicles which will enter and leave the road until time $t_0+\Delta \tfut$, 
the problem arises how to enforce these boundary conditions, which sensors clearly cannot provide at time $t$.
To do that, we set up a different LSTM-based ANN to predict sensor data in the time interval $[t_0,t_0+\Delta \tfut]$.
More precisely, the output of the ANN will not be the minute-by-minute flux data, since they are too fluctuating to guarantee a reliable prediction; instead, we opt to predict the \emph{total} number of vehicles in the time interval $[t_0,t_0+\Delta \tfut]$: a simpler yet useful piece of information since it can be interpreted as a constant boundary condition that, in most of the cases, offer a  good accuracy regarding the total mass found along the whole road at $t = t_0+\Delta\tfut$.

\subsection*{Paper organization}
The rest of the paper is organized as follows.
Section~\ref{sec:discuss-data} introduces the traffic flow data we considered and provides an overview of the benchmark dataset provided by Autovie Venete S.p.A.

Section~\ref{sec:enrich-data} introduces in general terms the structure of the ANN which will enrich the dataset.

Section~\ref{sec:flunas} discusses the training and the validation of the network which provides information about the real-time detection of congestion events. The same section also discusses short-term (few minutes) forecast of congestion events.

Section~\ref{sec:preconbo} discusses the training and the validation of the network which provides information about a mid-term (30 minutes) forecast of the expected traffic volume at sensors.

Section~\ref{sec:models} is devoted to the link between the enriched dataset obtained by using the ANNs previously described and the mathematical model, in the cases of both nowcast and forecast.

Finally, Section~\ref{sec:conclusions} concludes the work with some final remarks.

\section{Discussing the data}\label{sec:discuss-data}

At a macroscopic level, traffic flow is characterized by three state variables: the \emph{flux} $f$, corresponding to the number of vehicles passing through a given point per unit of time, the \emph{density} $\rho$, corresponding to the number of vehicles per unit of length, and the \emph{velocity} $v$. 
In the mathematical world, these quantities are functions defined on a continuous space-time domain and the following relation holds true 
\begin{equation}\label{f=rhov}
f(x,t)=\rho(x,t)\ v(x,t)
\end{equation}
for any point $x$ of the road and time $t$.
In the real world, however, these pieces of information can only be measured as discretized samples, therefore they cannot be calculated exactly at the same time.
It is also well known that computing $\rho$ by inverting \eqref{f=rhov} often leads to bad results, especially for large values of $\rho$, cf.\ \cite{herty2022pp}.

\begin{figure}
	\centering
	\includegraphics[width=.7\linewidth]{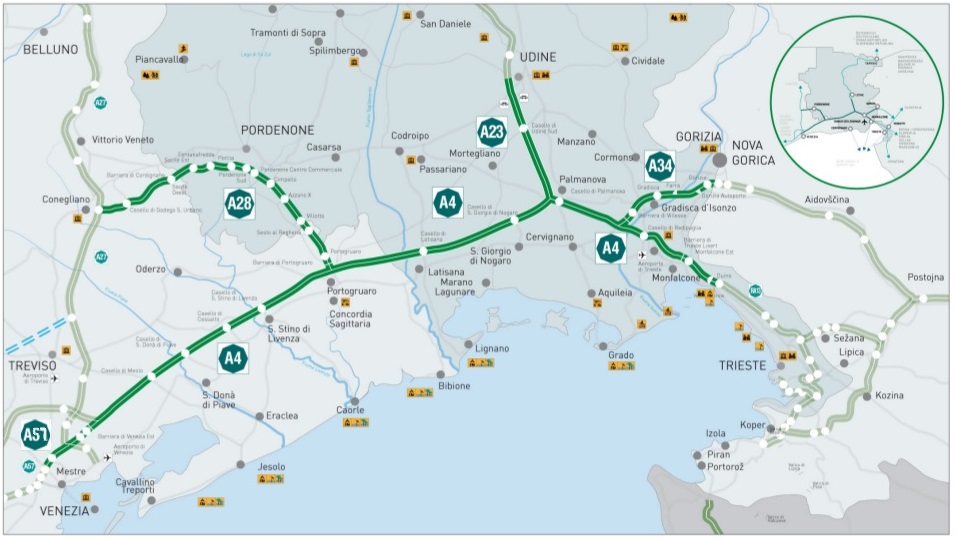}
	\caption{The Italian motorway A4 Trieste–Venice and its branches to/from Udine, Pordenone, and Gorizia, managed by Autovie Venete S.p.A.}
	\label{fig:cartina-autovie}
\end{figure}

Our benchmark dataset is provided by the Italian motorway company Autovie Venete S.p.A.\ and it was collected between September 2020 and March 2022. It contains traffic data from fixed sensors located along three highways in the North-East of Italy, namely A23, A28, and (part of the) A4, see Figure~\ref{fig:cartina-autovie}.
All of the highways have two lanes per direction except for the A4 which has three lanes in some parts. As usual, vehicles have different speed limits on the basis of their length and weight, moreover, heavy vehicles cannot use the fastest lane.

Data are collected by $45$ groups of sensors which provide, in total 301,200 records per day on average.
Each group is characterized by a position $x$ along the road and a direction of travel and consists of \emph{1 sensor per lane} (therefore we have 2 or 3 sensors per group).
Each sensor counts every vehicle that passes in front of it, along with its speed, and classifies it according to the German TLS
5+1 class standard \cite{TLS-class}.
The claimed error on counting is $\pm 3\%$ while the error on velocity is $\pm\max\{3\ \text{km/h}, 3\%\}$.
In the following, we aggregate classes 1 and 2 as \emph{light vehicles} (motorcycles, cars, vans, and car trailers) and classes 3, 4, and 5 as \emph{heavy vehicles} (lorries, lorry trailers, tractor vehicles and buses). 
In Sections \ref{sec:flunas} and \ref{sec:preconbo} we will further aggregate data by class and by group, respectively.

The spatial granularity is highly variable since the distance between sensors ranges from 1 to 20 km.
The temporal granularity is instead more regular since data are transmitted by each sensor every 1 minute, as \emph{aggregate measurements}: this means that the database stores the total number of vehicles passed in that time interval and the average velocity per each class. 
The measurements are kept for $2$ hours for real-time analysis then they are moved to a separate database. 
Consequently, historicized data are not available in real time. 

\begin{figure}[h!]
	\centering
	\includegraphics[width=.95\linewidth]{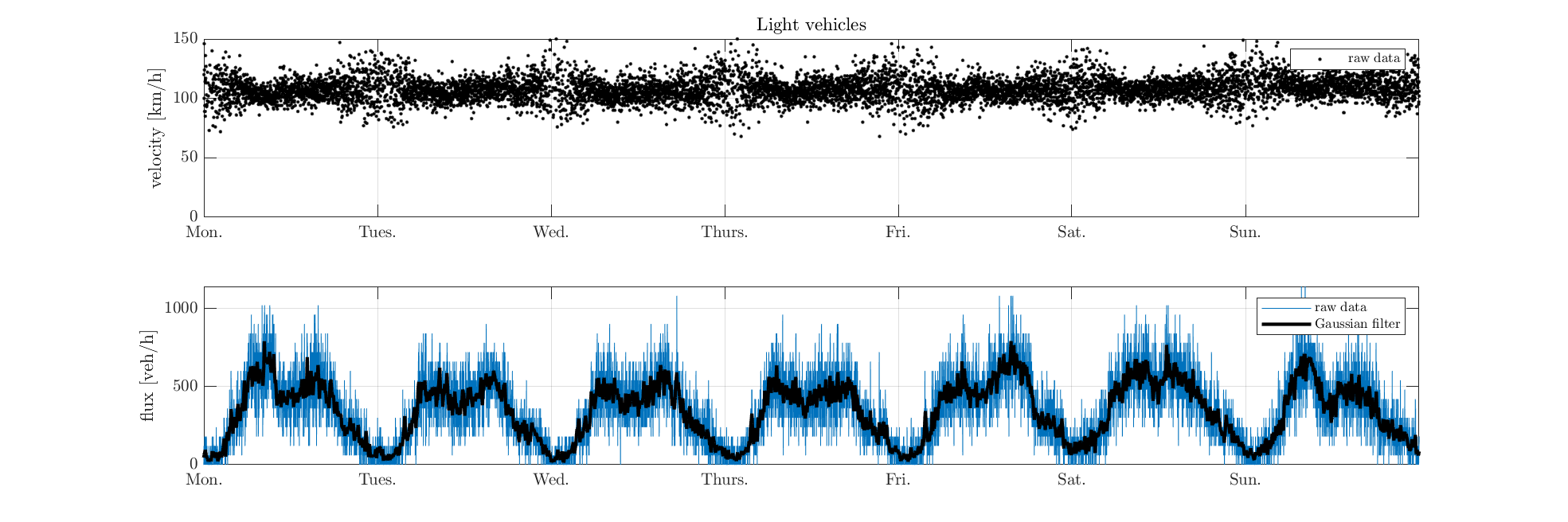}\\
	\includegraphics[width=.95\linewidth]{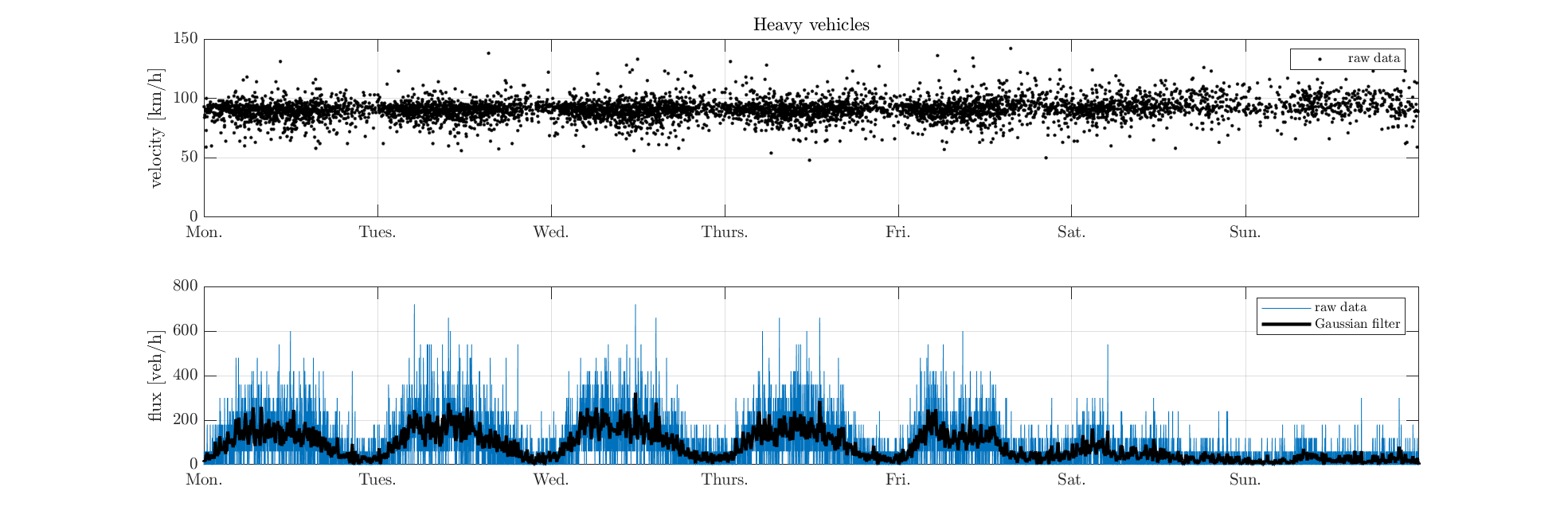}
	\caption{Velocity and flux for light (top) and heavy (bottom) vehicles, entire week from Monday to Sunday. We observe that the raw flux data are very fluctuating from minute to minute, but, applying a Gaussian filter (black line), one can recognize a certain pattern repeated on a daily basis. At night, the flux data of all vehicles drop, while the velocity data of light vehicles become more scattered. As expected, during the weekend, the flux of heavy vehicles is quite low.}
	\label{fig:datisettimanali}
\end{figure}

Some remarks are in order:
\begin{itemize}
	\item As mentioned above, large density values cannot be recovered by flux and velocity data simply inverting the relation \eqref{f=rhov}. The reason comes from a combination of the measurement aggregation and temporal granularity (low fluxes require long time intervals to be detected because vehicles move slowly).
	\item Although flux data show a regularity on a daily basis, they are very fluctuating from minute to minute, see Figure~\ref{fig:datisettimanali}.
	\item Most importantly, our data cannot distinguish between an empty road and a fully congested road. In both cases the measured flux is 0 and the velocity is undefined.
\end{itemize}

\begin{figure}[h!]
	\centering
	\includegraphics[width=0.99\linewidth]{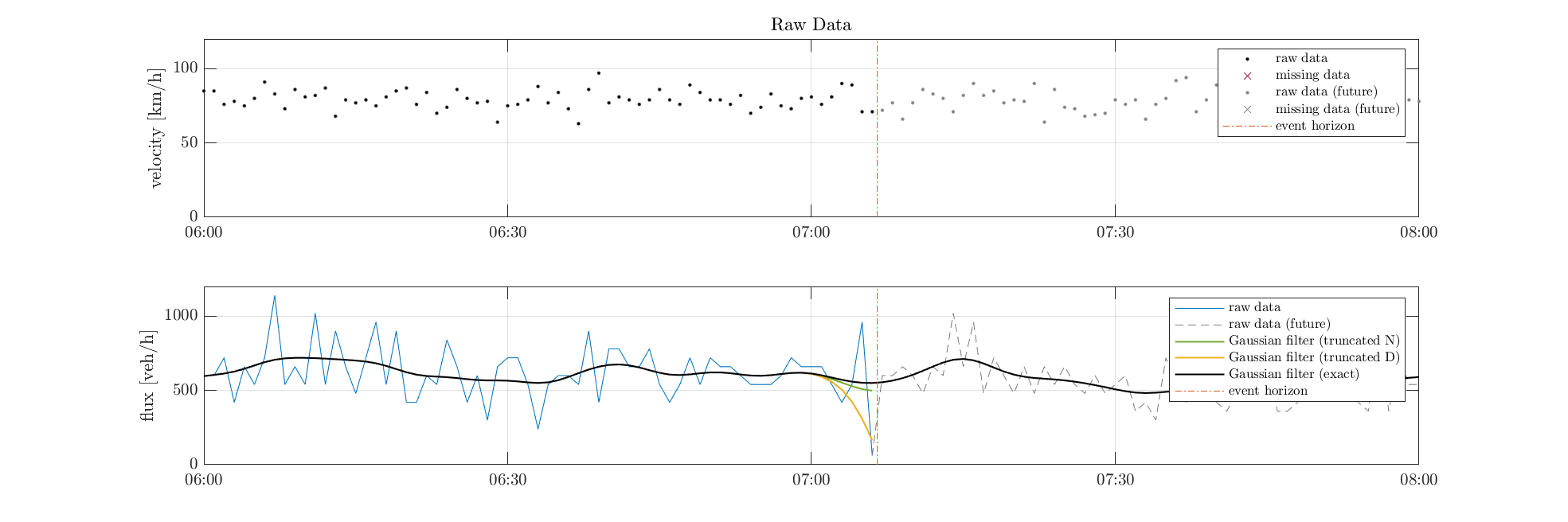}\\
	\includegraphics[width=0.99\linewidth]{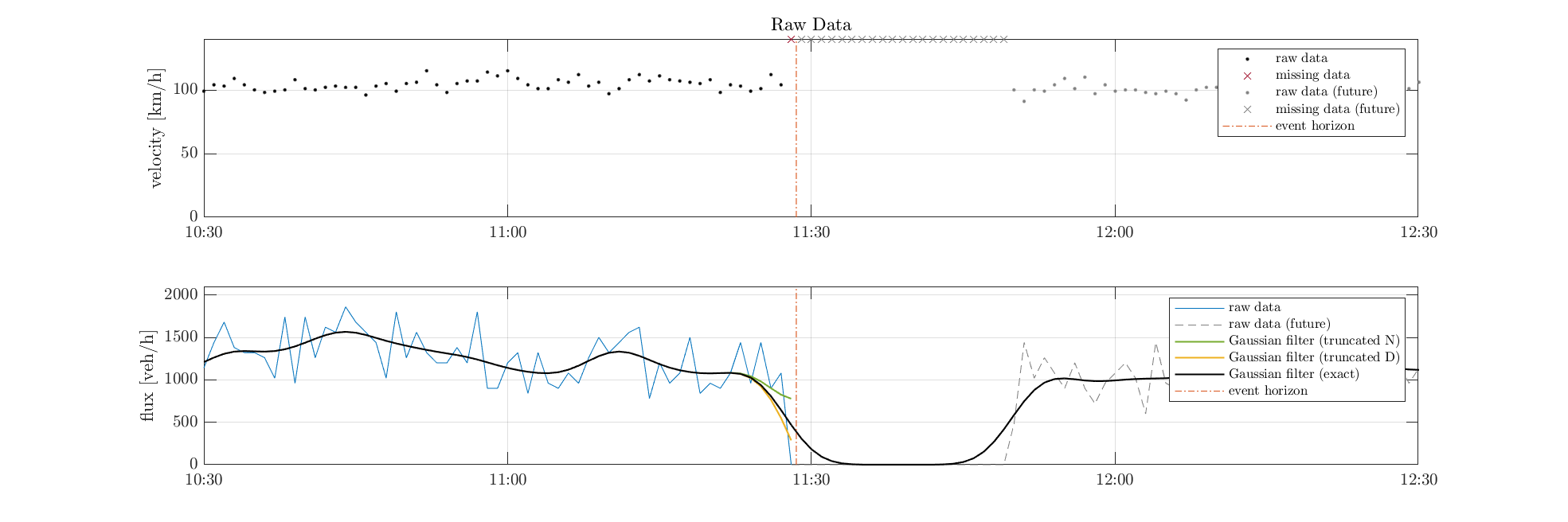}
	\caption{(top) Normal traffic conditions characterized by the usual high fluctuation of the flux. At 7:05 the flux drops abruptly then the  traffic restarts normally; (bottom) At 11:29 a very similar situation appears but, this time, it evolves into a queue. Beside the Gaussian filter already shown in Figure~\ref{fig:datisettimanali}, here we also show two other Gaussian filters obtained without using future data (beyond the event horizon). Truncation is obtained assuming either Dirichlet-like boundary conditions or Neumann-like boundary conditions. We see that neither Gaussian filters nor raw data are enough to distinguish between the two scenarios at the event horizon.}
	\label{fig:eventHorizon}
\end{figure}

These remarks are important to understand how it can be difficult to detect the formation of a congestion event in real time, which is, in turn, essential for a good estimation and forecast of the traffic flow, even far from the sensor which first observes the congestion.
In order to better understand this point, we show in Figure~\ref{fig:4conditions} four congestion events which develop with different characteristics.
In Figure~\ref{fig:eventHorizon}, instead, we show two very similar traffic conditions characterized by a flux drop, which evolve in a totally different manner: one into the free regime and the other into the congested regime.
This makes it clear that, despite it is relatively easy to detect congestion events (distinguishing them from empty road conditions) by observing data \emph{a posteriori} (e.g., a whole day), it is very difficult to do the same \emph{at the moment} of the congestion formation. 
{
	\def\widthscale{.49}
	\begin{figure}[h!]
		\centering
		\includegraphics[width=\widthscale\linewidth]{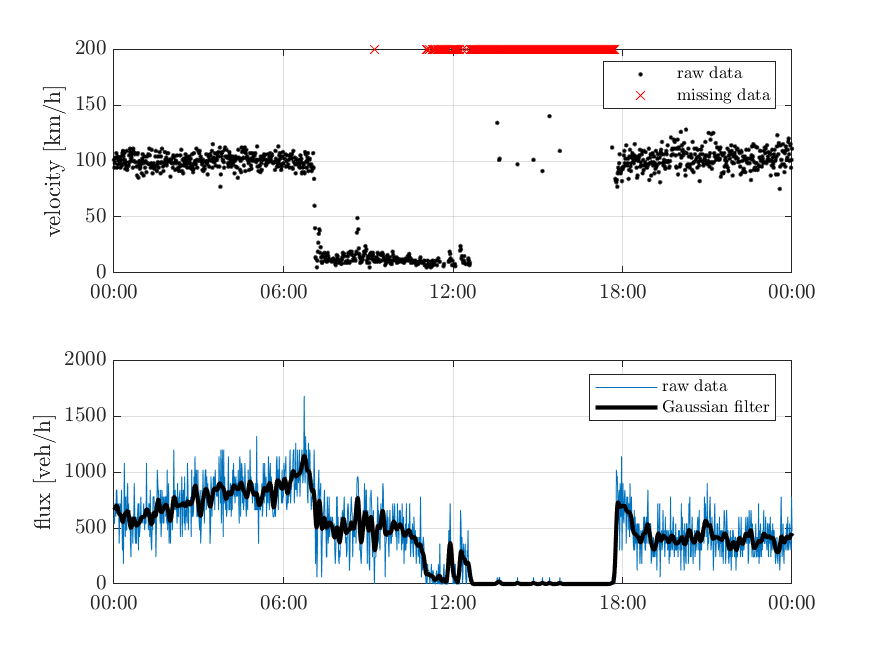}
		\includegraphics[width=\widthscale\linewidth]{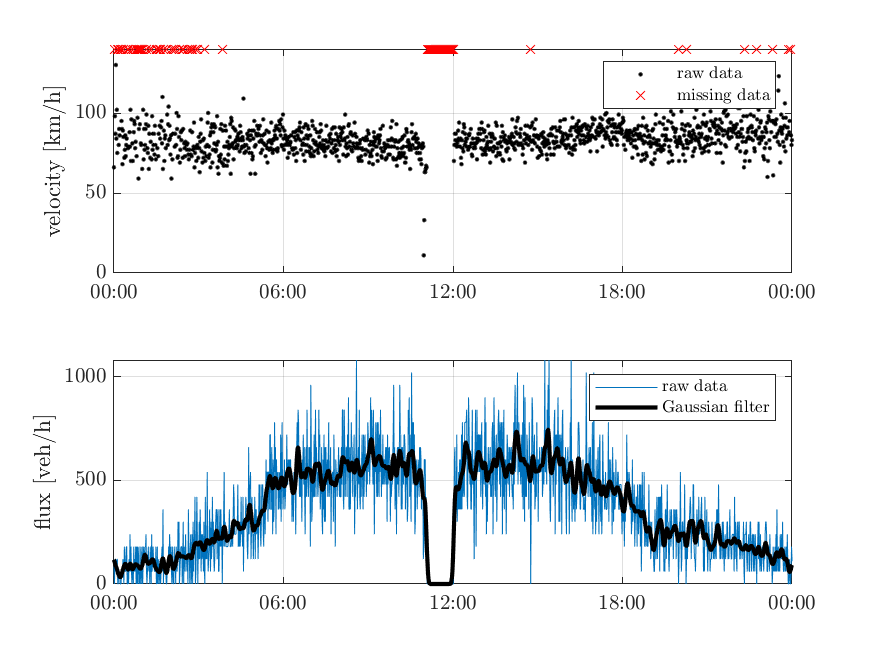}\\
        \includegraphics[width=\widthscale\linewidth]{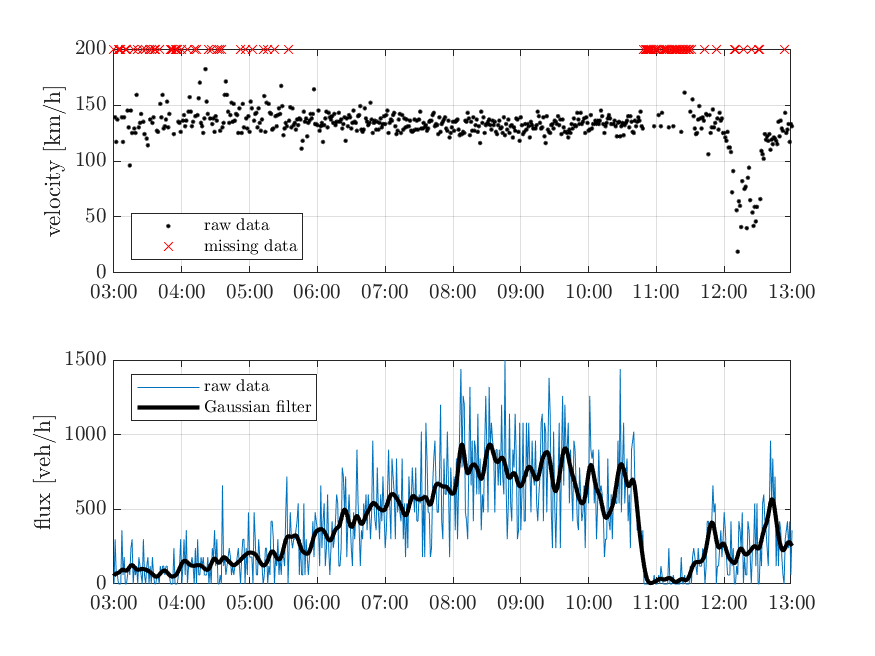}
		\includegraphics[width=\widthscale\linewidth]{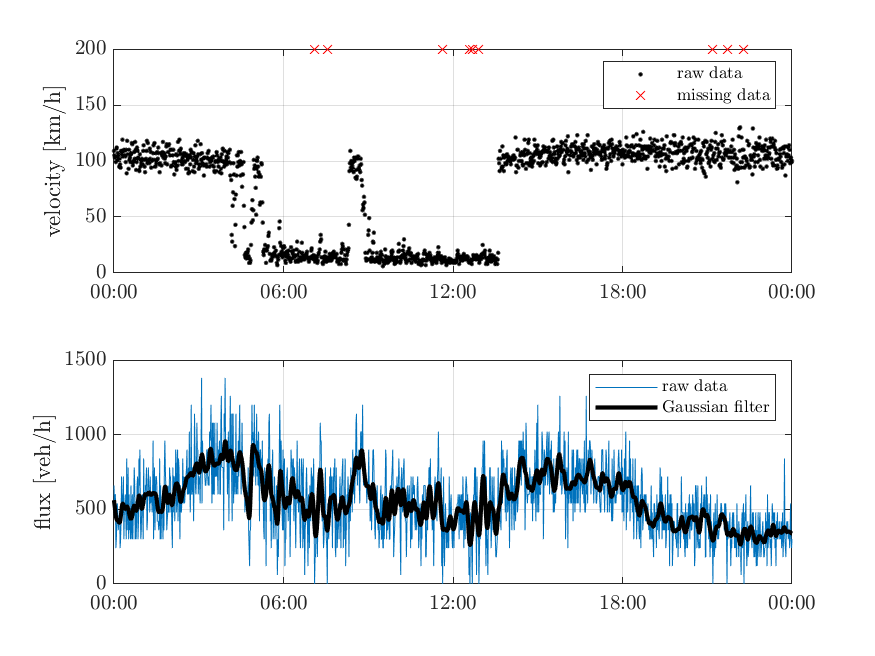}
		\caption{Four congestion events with different features: (top-left) we observe a rapid velocity drop and flux drop, then flux vanishes while velocity is undefined (with some exceptions for some fast vehicles still passing); (top-right) flux and velocity drop abruptly; (bottom-left) flux drops first, then velocity drops; (bottom-right) velocity drops while flux is only partially lowered.}
		\label{fig:4conditions}
	\end{figure}
}

For our purposes, it is crucial to note that $17$ groups of sensors out of the 45 deployed are also equipped with advanced technologies\footnote{Provided by \href{http://www.asim-technologies.com}{Asim Technologies Ltd}, series TT295.} 
that, combining Doppler radar, ultrasound emitters, and passive infrared radiation detectors, are able to determine also stopped vehicles and intense congestion conditions, reporting them as Boolean flags in the corresponding minute. 
In the following, we will refer to these sensors as \ASIM sensors and we will use them as one of the main sources of data for supervised learning purposes.

\clearpage
\section{Supervised machine learning approach for the dataset}\label{sec:enrich-data}

In this section, we set up a unique ML-based approach to solve both the two problems introduced in Section~\ref{sec:introduction}, namely the real-time detection of congestion events, and the forecasting of the expected average flux at sensors.

As we have already recalled, among the supervised learning techniques, RNN proved to be very effective in time series analysis. They are, in fact, a specific class of ANN devoted to analyzing temporal sequences $\vec{X} = (\vec{x}_t)_{t=1,2,\ldots}$ with a fixed number of features $\nin$ (namely $|\vec{x}_t| = \nin$), where the same computing unit is iteratively applied once per each step $t$ of the sequence taking as input the $\nin$ features of the current step along with the output of the previous step.
In our case, the features which can be extrapolated from the dataset are flux $f$ and velocity $v$ organized by class of vehicles and/or by lane.

In particular, LSTM-RNN is nowadays the preferred tool for many problems related to time series, being able to capture temporal features or dependencies also in long-range periods of time. 
The idea underlying the LSTM is to divide the output $\vec{h}_t$ (or \emph{hidden state}, in the LSTM jargon) of each step from the processed data that generates it, hence keeping a sort of ``internal memory'' $\vec{c}_t$ (or \emph{cell state}) of the LSTM itself. Both $\vec{h}_t$ and $\vec{c}_t$ shares the same dimension $\nhid$ (namely $|\vec{h}_t|=|\vec{c}_t| = \nhid$) which also represents the ``length'' of the memory (time window) the LSTM is capable to capture.
In particular, the output of the LSTM always relies only on the last $\nhid$ steps of the series under analysis: outputting the result after a number of steps smaller than $\nhid$ provides a less accurate answers.
Conversely to $\nin$, such parameter must be tuned with suitable approaches (hyper-tuning).
The cell state can be then updated depending on the (eventually normalized) novel input $\vec{x}_t$ conditioned over the previous output $\vec{h}_{t-1}$.

In practice, the LSTM update is based on four different components $\myvecw{g}{f}_t$, $\myvecw{g}{i}_t$, $\myvecw{g}{c}_t$, and $\myvecw{g}{o}_t$ (or \emph{gates}, with $|\myvec{g}{\cdot}| = \nhid$) that evaluate over independently weighted combinations of $\vec{h}_{t-1}$ and $\vec{x}_t$ as:
\begin{equation}
    \begin{split}
        \myvecw{g}{f}_t =&\ \text{sigmoid}\Big(\myvecw{b}{f} + \myvecw{W}{f} \times \vec{x}_t + \myvecw{R}{f} \times \vec{h}_{t-1} \Big),\\
        \myvecw{g}{i}_t =&\ \text{sigmoid}\Big(\myvecw{b}{i} + \myvecw{W}{i} \times \vec{x}_t + \myvecw{R}{i} \times \vec{h}_{t-1} \Big),\\
        \myvecw{g}{c}_t =&\ \text{tanh}\Big(\myvecw{b}{c} + \myvecw{W}{c} \times \vec{x}_t + \myvecw{R}{c} \times \vec{h}_{t-1} \Big),\\
        \myvecw{g}{o}_t =&\ \text{sigmoid}\Big(\myvecw{b}{o} + \myvecw{W}{o} \times \vec{x}_t + \myvecw{R}{o} \times \vec{h}_{t-1} \Big),\\
    \end{split}
\end{equation}
where $\times$ denotes the standard matrix-vector product, and $\myvec{W}{\cdot}$, $\myvec{R}{\cdot}$, and $\myvec{b}{\cdot}$ are called respectively \emph{input weights}, \emph{recurrent weights}, and \emph{biases}. Such weights are (typically) randomly assigned in the beginning and are the objective of the training phase through the back-propagation of the error.

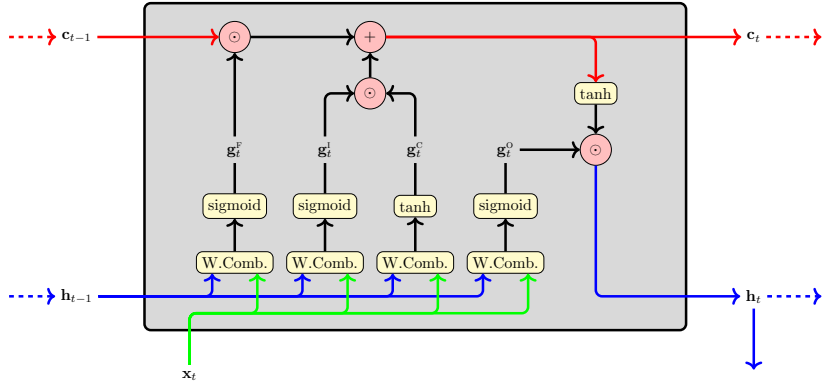
\begin{figure}
    \centering
    \scalebox{.6}{\def\funccolor{yellow!25}
\def\opcolor{red!25}
\def\boxcolor{black!15}

\begin{tikzpicture}[
    myvec/.style={},
    myfunc/.style={draw, rectangle, rounded corners, align=center, text centered, fill=\funccolor},
    myop/.style={draw, circle, align=center, text centered, fill=\opcolor},
    mybox/.style={ultra thick, rounded corners, fill=\boxcolor},
    myline/.style={->, ultra thick, rounded corners}
]

    \def\xlstm{0}
    \def\ylstm{0}
    \def\dx{2.0}
    \def\dy{1.25}
    \def\topborder{.75}
    \def\botborder{1.5}
    \def\outxborder{1.5}
    \def\outbotborder{1.0}
    \def\xhigh{6*\dx}
    \def\yhigh{\botborder+4*\dy+\topborder}

    \def\colx{green} 
    \def\colh{blue}  
    \def\colc{red}   
    \def\coll{black} 

    \draw[mybox] ($(\xlstm, \ylstm) + (0, 0)$) rectangle ++(\xhigh,\yhigh);

    \node[myvec] (ct0) at ($(\xlstm-\outxborder, \ylstm+\botborder+4*\dy)$) {$\vec{c}_{t-1}$};
    \node[myvec] (ht0) at ($(\xlstm-\outxborder, \ylstm+.5*\botborder)$) {$\vec{h}_{t-1}$};
    \node[myvec] (xt1) at ($(\xlstm+.5*\dx, \ylstm-\outbotborder)$) {$\vec{x}_{t}$};
    \coordinate[] (tmpx) at ($(\xlstm+\dx, \ylstm+.25*\botborder)$) {};

    \node[myvec] (ct1) at ($(\xlstm+\xhigh+\outxborder, \ylstm+\yhigh-\topborder)$) {$\vec{c}_{t}$};
    \node[myvec] (ht1) at ($(\xlstm+\xhigh+\outxborder, \ylstm+.5*\botborder)$) {$\vec{h}_{t}$};
    \node[myvec] (OUT) at ($(\xlstm+\xhigh+\outxborder, \ylstm-\outbotborder)$) {};

    \node[myfunc] (wcf) at ($(\xlstm+1*\dx, \ylstm+\botborder+0*\dy)$) {W.Comb.};
    \node[myfunc] (wci) at ($(\xlstm+2*\dx, \ylstm+\botborder+0*\dy)$) {W.Comb.};
    \node[myfunc] (wcc) at ($(\xlstm+3*\dx, \ylstm+\botborder+0*\dy)$) {W.Comb.};
    \node[myfunc] (wco) at ($(\xlstm+4*\dx, \ylstm+\botborder+0*\dy)$) {W.Comb.};

    \node[myfunc] (ff) at ($(\xlstm+1*\dx, \ylstm+\botborder+1*\dy)$) {sigmoid};
    \node[myfunc] (fi) at ($(\xlstm+2*\dx, \ylstm+\botborder+1*\dy)$) {sigmoid};
    \node[myfunc] (fc) at ($(\xlstm+3*\dx, \ylstm+\botborder+1*\dy)$) {tanh};
    \node[myfunc] (fo) at ($(\xlstm+4*\dx, \ylstm+\botborder+1*\dy)$) {sigmoid};
    \node[myfunc] (fh) at ($(\xlstm+5*\dx, \ylstm+\botborder+3*\dy)$) {tanh};

    \node[myvec] (gf) at ($(\xlstm+1*\dx, \ylstm+\botborder+2*\dy)$) {$\myvecw{g}{f}_t$};
    \node[myvec] (gi) at ($(\xlstm+2*\dx, \ylstm+\botborder+2*\dy)$) {$\myvecw{g}{i}_t$};
    \node[myvec] (gc) at ($(\xlstm+3*\dx, \ylstm+\botborder+2*\dy)$) {$\myvecw{g}{c}_t$};
    \node[myvec] (go) at ($(\xlstm+4*\dx, \ylstm+\botborder+2*\dy)$) {$\myvecw{g}{o}_t$};

    \node[myop] (mul1) at ($(\xlstm+1*\dx, \ylstm+\botborder+4*\dy)$) {$\odot$};
    \node[myop] (mul2) at ($(\xlstm+2.5*\dx, \ylstm+\botborder+3*\dy)$) {$\odot$};
    \node[myop] (mul3) at ($(\xlstm+5*\dx, \ylstm+\botborder+2*\dy)$) {$\odot$};
    \node[myop] (sum1) at ($(\xlstm+2.5*\dx, \ylstm+\botborder+4*\dy)$) {$+$};

    \draw[myline, color=\colc] (ct0) -- (mul1);
    \draw[myline, color=\colc] (sum1) -- (ct1);
    \draw[myline, color=\colc] (sum1) -| (fh);

    \draw[myline, color=\colh] (ht0) -| ($(wcf.south)-(.5,0)$);
    \draw[myline, color=\colh] (ht0) -| ($(wci.south)-(.5,0)$);
    \draw[myline, color=\colh] (ht0) -| ($(wcc.south)-(.5,0)$);
    \draw[myline, color=\colh] (ht0) -| ($(wco.south)-(.5,0)$);
    \draw[myline, color=\colh] (mul3) |- (ht1);
    \draw[myline, color=\colh] (ht1) -- (OUT);

    \draw[myline, color=\colx] (xt1) |- (tmpx) -| ($(wcf.south)+(.5,0)$);
    \draw[myline, color=\colx] (xt1) |- (tmpx) -| ($(wci.south)+(.5,0)$);
    \draw[myline, color=\colx] (xt1) |- (tmpx) -| ($(wcc.south)+(.5,0)$);
    \draw[myline, color=\colx] (xt1) |- (tmpx) -| ($(wco.south)+(.5,0)$);

    \draw[myline, color=\coll] (wcf) -- (ff);
    \draw[myline, color=\coll] (wci) -- (fi);
    \draw[myline, color=\coll] (wcc) -- (fc);
    \draw[myline, color=\coll] (wco) -- (fo);

    \draw[myline, color=\coll] (ff) -- (gf) -- (mul1);
    \draw[myline, color=\coll] (fi) -- (gi) |- (mul2);
    \draw[myline, color=\coll] (fc) -- (gc) |- (mul2);
    \draw[myline, color=\coll] (fo) -- (go) -- (mul3);

    \draw[myline, color=\coll] (mul1) -- (sum1);
    \draw[myline, color=\coll] (mul2) -- (sum1);
    \draw[myline, color=\coll] (fh) -- (mul3);

    \draw[myline, color=\colc, dashed] ($(ct0)-(\outxborder, 0)$) -- (ct0);
    \draw[myline, color=\colc, dashed] (ct1) -- ($(ct1)+(\outxborder, 0)$);
    \draw[myline, color=\colh, dashed] ($(ht0)-(\outxborder, 0)$) -- (ht0);
    \draw[myline, color=\colh, dashed] (ht1) -- ($(ht1)+(\outxborder, 0)$);
\end{tikzpicture}}
    \caption{
        Schematic structure of the LSTM computing unit evaluating input at time $t$.
        W.Comb.\ represents the weighted combination with bias given as $\myvec{b}{\cdot} + \myvec{W}{\cdot} \times \vec{x}_t + \myvec{R}{\cdot} \times \vec{h}_{t-1}$.
        All the components are vectors of length $\nhid$ but $\vec{x}$ which is of length $\nin$.
        We recall that $\vec{h}_t$ is also the output of the network.
    }
    \label{fig:lstm}
\end{figure}

Each gate serves a different purpose, where the first three rule the update of the cell state $\vec{c}$ and the fourth determine the next hidden state $\vec{h}$ (regulating the cell state contribution). More formally:
\begin{enumerate}
    \item[$\myvecw{g}{f}$] or \textit{forget gate:} weakens $\vec{c}_{t-1}$ by applying a transformation within the range $(0,1)$.
    \item[$\myvecw{g}{i}$] or \textit{input gate:} decides how the candidate influences $\vec{c}_t$, being a transformation of range $(0,1)$.
    \item[$\myvecw{g}{c}$] or \textit{candidate gate:} represents the cell input activation that regulates $c$, being of range $(-1,1)$.
    \item[$\myvecw{g}{o}$] or \textit{output gate:} decides how $\vec{c}_t$ will compose the output, applying a transformation of range $(0,1)$.
\end{enumerate}
Hence, (see also Figure~\ref{fig:lstm} for a pictorial representation) the update rules are given by
\begin{equation}
    \vec{c}_{t} = \myvecw{g}{f}_t \odot \vec{c}_{t-1} + \myvecw{g}{i}_t \odot \myvecw{g}{c}_t,
    \qquad \qquad
    \vec{h}_t = \myvecw{g}{o}_t \odot \tanh(\vec{c}_t),
\end{equation}
where $\odot$ denotes the Hadamard (element-wise) product.

In particular, it is important to notice that the output $\vec{h}$ of the LSTM is a $\nhid$ vector and, consequently, it needs to be manipulated to either give a prediction or a classification.
Our tool of choice is a vanilla SLP Feed-Forward Network that condenses the $\nhid$ features in 
\begin{itemize}
    \item[-] \emph{Prediction task}: a $\npred$-lenght output $\vec{o}$, where each entry represents an individual prediction.
    \item[-] \emph{Classification task}: a $\nclass$-length features-vector $\vec{z}$, where each entry represents a class of the problem, then fed into a softmax layer to transform them into a probability vector $\hat{\vec{o}}$ (of being of a specific class).
    We recall the softmax function is defined component-wise as follows:
    \begin{equation}
        \text{softmax}(\vec z)_i = \frac{e^{z_i}}{\sum_{j=0}^{\nclass-1} e^{z_j}}, \qquad i=0, \dots, \nclass-1.
    \end{equation}
    The final output can be either the probability vector $\hat{\vec{o}}$ or the index of the highest-probability class $o$, \ie $o = \text{argmax}(\hat{\vec{o}})$. Do note that $\hat{\vec{o}}$ can be used as a \emph{confidence} indicator of the prediction as the closer is $\max(\hat{\vec{o}})$ to $1$, the more certain is the prediction according to the ANN.
\end{itemize}
We recall that SLP is a regular weighted combination with bias, namely the output $\vec{z}$ of length either $\npred$ or $\nclass$ is obtained as $\vec{z}_t = \vec{b} + \vec{W} \times \vec{h}_t$, where $\vec{W}$ and $\vec{b}$ are a suitable-sized weight-matrix and bias-vector.

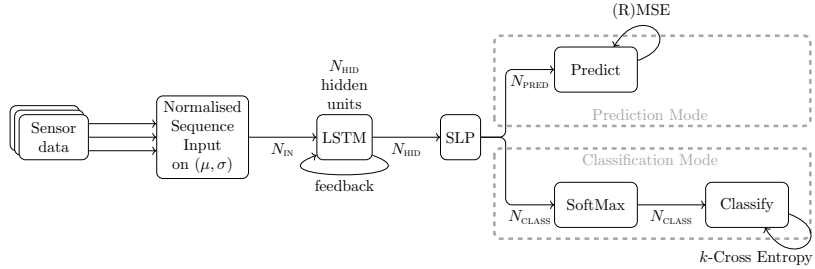
\begin{figure}[h!]
	\centering
	\scalebox{.6}{\def\bgcolor{white}
\def\boxcolor{black!35}

\begin{tikzpicture}[
    mynode/.style={draw, rectangle, rounded corners, align=center, text centered, minimum height=1cm, fill=\bgcolor},
    mybox/.style={dashed, ultra thick, rounded corners, color=\boxcolor},
    node distance=1.5cm
]
    \def\dy{1.5}
    
    \node   () [mynode, text width=1.3cm] at (-.2,+.2) {};
    \node  (0) [mynode, text width=1.3cm, label={[yshift=.2cm]}] at (-.1,+.1) {};
    \node  (1) [mynode, text width=1.3cm] at (0, 0) {Sensor data};
    \node  (2) [right=of 1, mynode, text width=1.8cm] {Normalised Sequence Input\\ on $(\mu, \sigma)$};
    \node  (3) [right=of 2, mynode, label={[align=center, text width=1.5cm]$\nhid$ hidden units}] {LSTM};
    \node  (4) [right=of 3, mynode] {SLP};
    \node (5a) [mynode, text width=1.6cm] at ($(4)+(3,+\dy)$) {Predict};
    \node (5b) [mynode, text width=1.6cm] at ($(4)+(3,-\dy)$) {SoftMax};
    \node (6b) [right=of 5b, mynode, text width=1.6cm] {Classify};
    
    \draw[mybox] ($(4)+(.75,.25)$) rectangle ++(7,2);
    \draw[mybox] ($(4)+(.75,-.25)$) rectangle ++(7,-2);
    
    \node[color=\boxcolor] at ($(4)+(.7,.5)+(3.5,0)$) {Prediction Mode};
    \node[color=\boxcolor] at ($(4)+(.7,-.5)+(3.5,0)$) {Classification Mode};

    \draw[->] (1) -- (2);
    \draw[->] ([yshift=+.3cm]1.east) -- ([yshift=+.3cm]2.west);
    \draw[->] ([yshift=-.3cm]1.east) -- ([yshift=-.3cm]2.west);
    \draw[->] (2) -- node[below]{$\nin$} (3);
    \draw[->] (3) -- node[below]{$\nhid$} (4);
    \draw[->, rounded corners] (4) -- ($(4)+(1,0)$) -- ($(4)+(1,+\dy)$) -- node[below]{$\npred$} (5a);
    \draw[->, rounded corners] (4) -- ($(4)+(1,0)$) -- ($(4)+(1,-\dy)$) -- node[below]{$\nclass$} (5b);
    \draw[->] (5b) -- node[below]{$\nclass$} (6b);
    \draw[->] ([yshift=-.4cm]3.east) to[out=-20,in=-160,looseness=3.5] node[below]{feedback} ([yshift=-.4cm]3.west);
    \draw[->] ([yshift=+.2cm]5a.east) to[out=+20,in=+60,looseness=5] node[above,xshift=-.3cm,yshift=.1cm]{(R)MSE} ([xshift=+.4cm]5a.north);
    \draw[->] ([yshift=-.2cm]6b.east) to[out=-20,in=-60,looseness=5] node[below,xshift=-1.1cm]{$k$-Cross Entropy} ([xshift=+.4cm]6b.south);
\end{tikzpicture}}
	\caption{
		Pipeline of the enriching tool.
		The sensor data sequence is globally feature-wise normalized before feeding it in the LSTM, one step at a time.
		The output of the LSTM is then processed by an MLP Network having either a single or no hidden layer (actually resulting in a single convolution step).
		Finally, depending whether it is a classification task or not, a SoftMax layer is applied before the result is given.
	}
	\label{fig:methodology}
\end{figure}

To complete the pipeline of our methodology (a summary can be found in Figure~\ref{fig:methodology}), we further need to specify how to evaluate the system performance against our ground-truth data, i.e.\ we need to choose an error function to evaluate the distance of our guesses from ground truth (a loss function, in the jargon).
\begin{itemize}
    \item[-] \emph{Prediction task}: being the prediction output a real number, since we are not interested in weighting differently over- and under-errors, we went for the vanilla (Root) Mean Squared Error.
    Denoting with $\vec{y}_t$ and $\vec{o}_t$ respectively the ground truth and the output of the network corresponding to the input $\vec{x}_t$, we recall the (root) mean squared error being:
    \begin{equation}\label{eq:mse}
        \text{MSE} = \frac1{|\vec{X}|}\frac{1}{\npred}\sum_{t=1}^{|\vec{X}|}\|\vec{o}_t - \vec{y}_t\|^2 \ , \qquad\qquad \text{RMSE} = \sqrt{\text{MSE}} \ .
    \end{equation}
    
    \item[-] \emph{Classification task}: a very common choice for Boolean classification loss is the $k$-cross entropy (with $k=2$).
    In particular, since our dataset shows a great disequilibrium between positive and negative samples, we decide (instead of enriching it with synthetic data that are often complex to generate) to weight the loss entropy on the positive ratio $0 < p_r < 1$.
    We recall that the softmax applied to a binary classification task returns a 2-value vector $\hat{\vec{o}}_t = (1-\hat{o}_t, \hat{o}_t)$ where $\hat{o}_t$ represents the probability that the positive class is chosen. Hence, if $y_t$ is the ground-truth bit corresponding to input $\vec{x}_t$ -- meaning that our target $\vec{y}_t$ is either $(1, 0)$ or $(0, 1)$ -- then the weighted binary-cross entropy is defined as:
        \begin{equation}
            \text{CrossEntropy}=-\frac1{|\vec{X}|}\sum_{t=1}^{|\vec{X}|}\bigg(\big((1-p_r) \cdot y_t \cdot \ln(\hat{o}_t) \big) + \big(p_r \cdot (1 - y_t) \cdot \ln(1 - \hat{o}_t)\big)\bigg)\ .
        \end{equation}
\end{itemize}

\section{Detection and short-term forecast of congestion}\label{sec:flunas}

In this Section, we build a labeled dataset of congestion events and we use the previously described ANN-based methodology to build two classifiers: a first classifier $\FFFF_c$ for performing real-time detection of congestion events, and a second classifier $\FFFF_p$ to perform short-term forecasting of the same congestion events. The two classifiers act as a sort of ``congestion alarm'' and ``congestion pre-alarm'' launchers, respectively.

\subsection{Building the dataset}

The dataset is created using data gathered by the \ASIM sensors. Since the congestion event is detected by each sensor (i.e., in each lane), we decided to aggregate the data of all classes of vehicles and working lane by lane. 
Therefore, the features are the total flux $f$ and the average velocity $v$ of all vehicles (all classes).
(Actually, recovering \emph{a posteriori} a piece of information per class is often quite easy. In fact, if a congestion event is detected in the slow lanes and not in the fast lane, it is highly probable that the congestion is for heavy vehicles only, since they cannot use the fast lane, see Section \ref{sec:discuss-data}.)

We split the data per day, so to perform an incremental training phase using only randomized batches of days of data at a given time. Therefore, we get dataset samples constituted by a two-feature 1440-long (24 h $\times$ 60 min) sequence $\vec{X} = (\vec{x}_1, \dots, \vec{x}_{1440})$, with $\vec{x}_t = (f_t, v_t)$, $t=1,\ldots, 1440$. The dataset should be labeled with a suitable 1440-long Boolean array $\vec{y} = (y_1, \dots, y_{1440})$ reporting whether the corresponding minute $t$ is labeled as a congestion event or not.
The array $\vec y$ is computed by means of a logical combination of three different computational procedures, each of which leads to a Boolean flag. More precisely, we have.
$$
\vec y=\vec{b}^{3T}\ \lor \ \vec{b}^f\ \lor \ \vec{b}^v,
$$
where $\lor$ denotes the logical \textit{or} and:
\begin{itemize}
\item 
$\vec{b}^{3T}$ is the flag for the congestion event transmitted directly by the \ASIM sensors.
On average, the amount of daily congestion events per sensor is pretty low ($<0.1\%$), hence, in order to make the dataset more balanced, we selected the samples reporting at least $\sim 1\%$ of positive labeling (\ie $\sum_{t=1}^{1440} ({b}^{3T}_t\geq 15$).
We noticed that the \ASIM detection procedure is quite ``conservative'', meaning that it tends to report a congestion event only if there is a stable queue under the sensor.
Moreover, the exact definition of congestion event, as well as the physical and computational procedure used to detect it are not publicly available and they are not known by the authors.
For these reasons, we decided to enrich the true-sample with the two following heuristics.
\item 
\MLN considers the flux array $\mathbf f:=(f_1, \dots, f_{1440})$ only, along with a 10-fold Gaussian regularization $\mathbf{f}^*$ obtained via a convolution with a triangular kernel, cf.\ Figure~\ref{fig:datisettimanali};
a congestion event is reported at time $t$ if the sensor detects the following condition during the daytime (from 5 AM to 8 PM):
\begin{equation}
    f_{t-1} < 2, \qquad\land\qquad f_t < 2, \qquad\land\qquad f^*_t < 2, \qquad\land\qquad \left(\frac{1}{60}\sum_{s = t-60}^{t-1} f_s\right)-f^*_t > 2 \ ,
\end{equation}
where $\land$ denotes the logical \emph{and} operator.
The idea behind this heuristic is that a sufficient condition for a congestion event is a low flux at the current time combined with a sudden drop of the flux (\ie, a high average flux in the previous time period). 
\item 
\NDR considers instead the velocity array $\mathbf v:=(v_1, \dots, v_{1440})$ only, along with a regularization $\vec{\mathbf v^*}$ (obtained as before). A congestion event is reported at time $t$ if the sensor detects the following conditions during the daytime (from 5 AM to 8 PM):
\begin{equation}
    v_t < v_{t-1}, \qquad\land\qquad 0 < v_t < 65, \qquad\land\qquad \max\{v^*_s - v_t \mid s = t-1, \dots, t-15\} \geq 40\ .
\end{equation}
The idea behind this heuristic is that a sufficient condition for a congestion event is a low velocity at the current time combined with a sudden drop of the velocity (\ie, a high velocity in the previous time period). 
\end{itemize}
\textbf{Remark.} The two heuristics were developed under a collaboration with the data owner and have been human-validated through over a month of direct observations.
\\
\textbf{Remark.} We point out that the two heuristics both use data that are not available yet at time $t$ (because of the regularizations), making them useless to perform real-time applications. 

\medskip

Analyzing the average and the standard deviation of the flux data we found that most of the sensors dispatched on A4 behave very differently from the others ($\mu \sim 1\pow+4$ vs.\ $2\pow+3$ vehicles/day and $\sigma \sim 1\pow+4$ vs.\ $5\pow+3$).
This is not surprising since A4 connects more populated areas compared to A23 and A28.
For this reason, we split the sensors into two disjoint sets, namely \emph{high flux} (HF) and \emph{low flux} (LF), being LF all the sensors in A23, A28, and in the fastest lane of A4 (where it has 3 lanes).
In the rest of this and the following section we consider HF sensors only, the procedure and analysis for LF sensors being similar.

We split the HF 119-day dataset $\{(\myvecp{X}{d},\myvecp{y}{d})\}$, $d=1,\ldots,119$ (171,360 training minutes) in $96$ training days and $23$ validation days.
The average sample positive rate after the enrichment is $p_r = 4.2\%$, hence we weighted the $k$-cross entropy function opportunely.

\subsection{Training the model}

We started by training the HF classifier $\FFFF_c$ first: we performed multiple training sessions to estimate the suitable size $\nhid$ of the memory of the LSTM.
Each training session was carried out by using the ADAM optimizer working on randomly shuffled batches of size 24 samples over 100 epochs arranged with 9 progressive refinements of the learning rate (10 epochs per refinement).

\begin{figure}
    \centering
    \includegraphics[width=.47\linewidth]{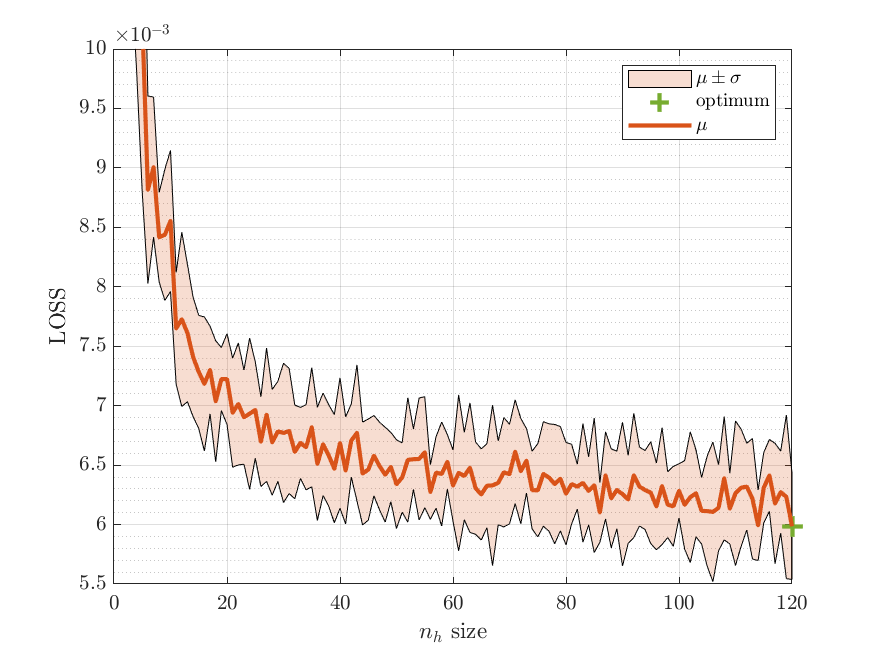}\hfill
    \includegraphics[width=.47\linewidth]{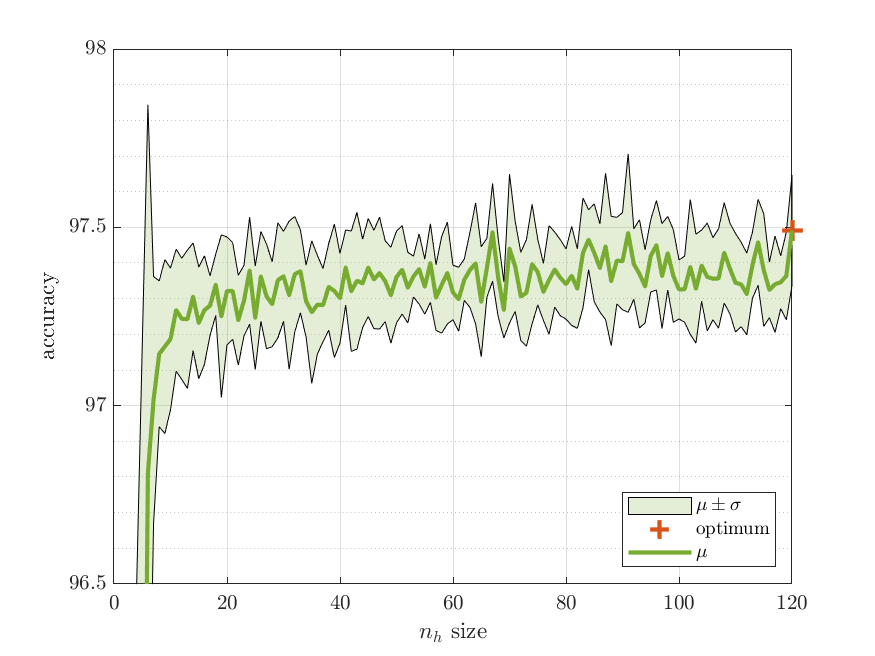}
    \caption{$k$-Cross Entropy (loss, on the left) and accuracy (on the right) achieved by the training sessions for $\FFFF_c$ at the varying of $\nhid$. As it can be seen, the better parameter is $\nhid = 120$ (i.e.\ corresponding to 2 hours) suggesting that having more data would refine the results even more.}
    \label{fig:kCrossEntropyNh}
\end{figure}

Figure~\ref{fig:kCrossEntropyNh} reports the results in terms of loss and accuracy for the various training sessions.
We found $\nhid = 120$ being the most suitable parameters for the model (with $\nin = 2$, $\nclass=2$), yielding best accuracy of $97.70\%$ ($99.75\%$ if weighted w.r.t.\ $p_r$) with corresponding loss value $5.2\times10^{-3}$. 
We recall that $\nhid = 120$ was the largest amount of data available in real-time applications for our framework, see Section \ref{sec:discuss-data}.

\begin{figure}
    \centering
    \includegraphics[width=.47\linewidth]{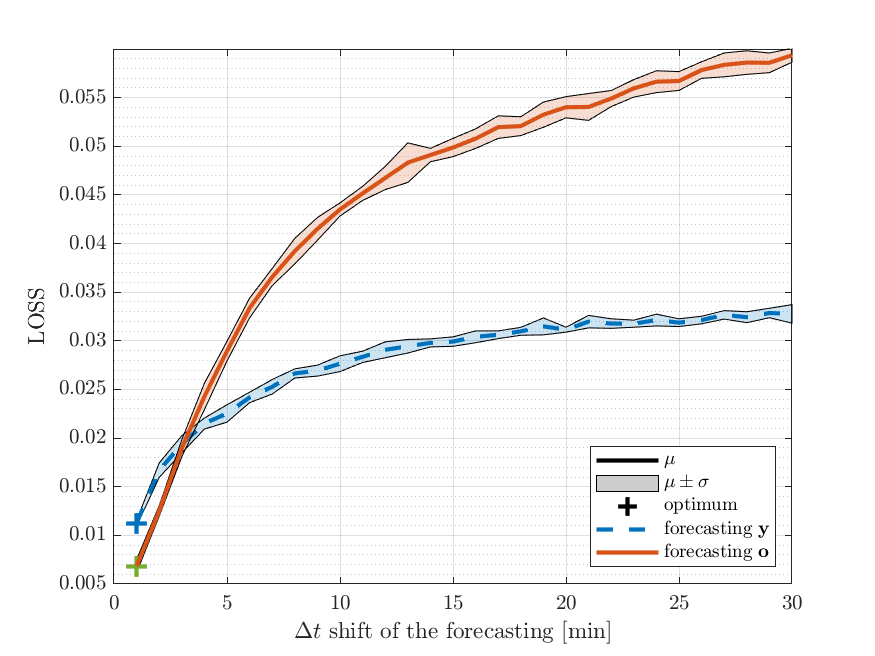}
    \hfill
    \includegraphics[width=.47\linewidth]{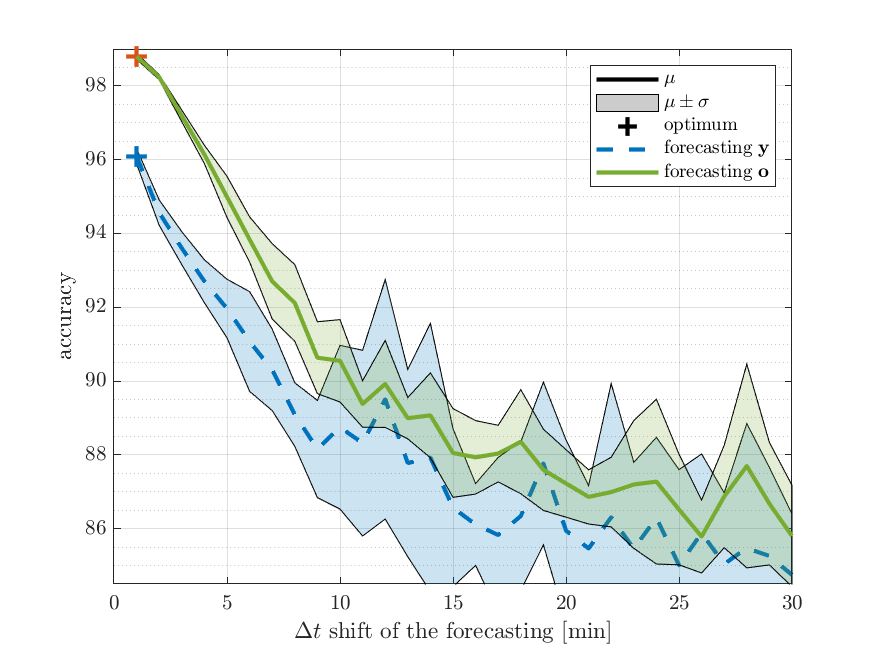}
    \caption{
        $k$-Cross Entropy (loss, on the left) and accuracy (on the right) achieved by the training sessions for $\FFFF_p$ at the varying of the time window of the forecasting $\Delta t$ ($x$-axis) both for predicting $\myvecp{y}{d}$ (in dashed blue) and $\myvecp{o}{d}$ (in orange/green).
        As expected, the smaller $\Delta t$, the better the prediction.
    }
    \label{fig:kCrossEntropyDelta}
\end{figure}

\medskip

Secondly, we tackled the problem of training the classifier $\FFFF_p$. 
We fixed the dimensions of the model -- namely $\nin = 2$, $\nhid = 120$, and $\nclass = 2$ -- and we focused on how many minutes we could anticipate the classification of $\FFFF_c$ without losing too much in accuracy.
The trivial way would be producing a labeled dataset for the forecasting task by applying a $\Delta t$ minutes shift to the vectors $\myvecp{y}{d}$ in order to ``bring backward'' the congestion events.
However, we prefer forecasting the output $\myvecp{o}{d}$ of $\FFFF_c$ itself (\ie the argmax of the probability vector) rather than the target $\myvecp{y}{d}$ of $\FFFF_c$. 
This change in objective is in fact convenient for two main reasons: (i) the output $\myvecp{o}{d}$ is someway a ``regularized'' version of the real target $\myvecp{y}{d}$, hence it should be easier to forecast; (ii) philosophically speaking, we are expecting that a $\FFFF_p$ alarm to be followed by a $\FFFF_c$ one.
In order to further correct the dataset, however, we also filtered the novel target vector from all the isolated congestion events, actually filtering the ones that probably were false-positives of $\FFFF_c$.

The results of the training with the two proposed datasets are reported in Figure~\ref{fig:kCrossEntropyDelta}, where loss and accuracy are compared at the varying of the $\Delta t$ shift.
Analyzing the accuracy, it can be seen that trying to forecast $\myvecp{o}{d}$ is in general simpler; the loss, on the contrary, grows slower when the target is set to $\myvecp{y}{d}$ actually prompting that training on $\myvecp{o}{d}$ is less conservative in terms of false-positive. 
We found $\Delta t = 4$ min being the most suitable parameter (in particular with the second dataset, still having comparable loss) yielding interesting predictions while keeping the average accuracy above $95\%$ (with the best peak of $96.50\%$, or $98.82\%$ if weighted) with a corresponding average loss value of $0.24$, hence comparable to results obtained by $\FFFF_c$.

\subsection[Performance evaluation of the real-time classifier]{Performance evaluation of the real-time classifier $\FFFF_c$}

{
    \def\widthscale{.49}

    \begin{figure}
        \centering
        \includegraphics[width=\widthscale\linewidth]{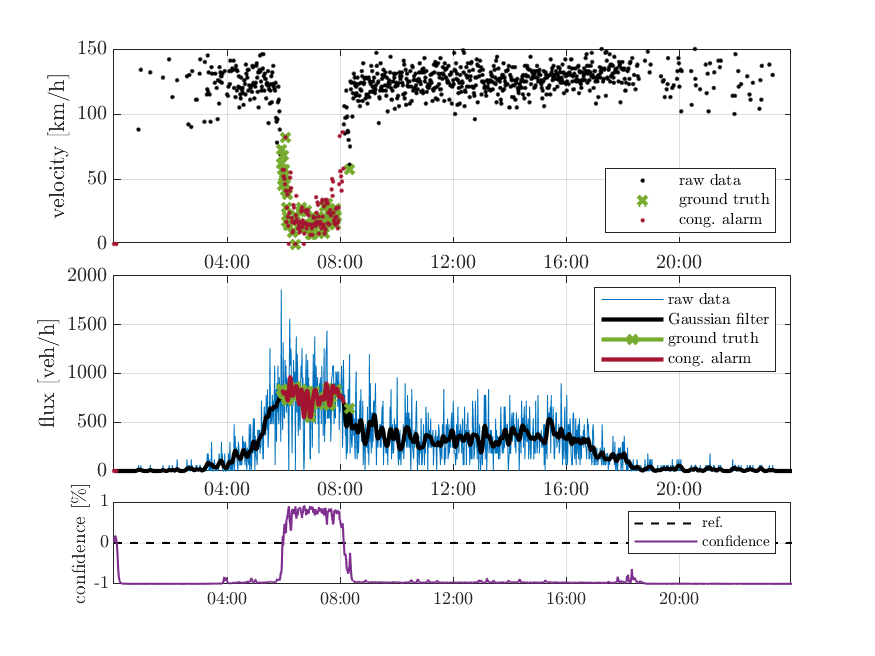}
        \includegraphics[width=\widthscale\linewidth]{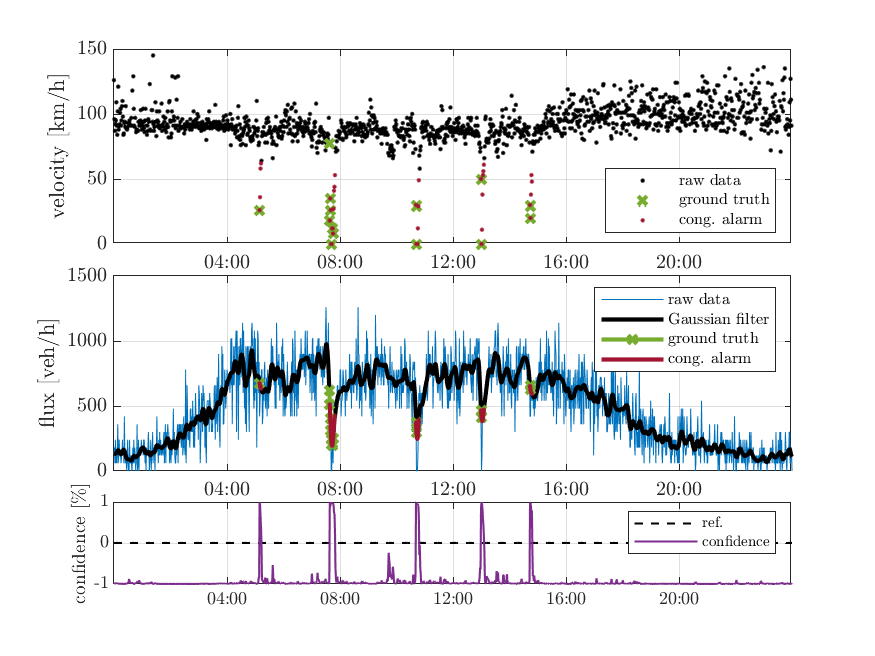}\\
        \includegraphics[width=\widthscale\linewidth]{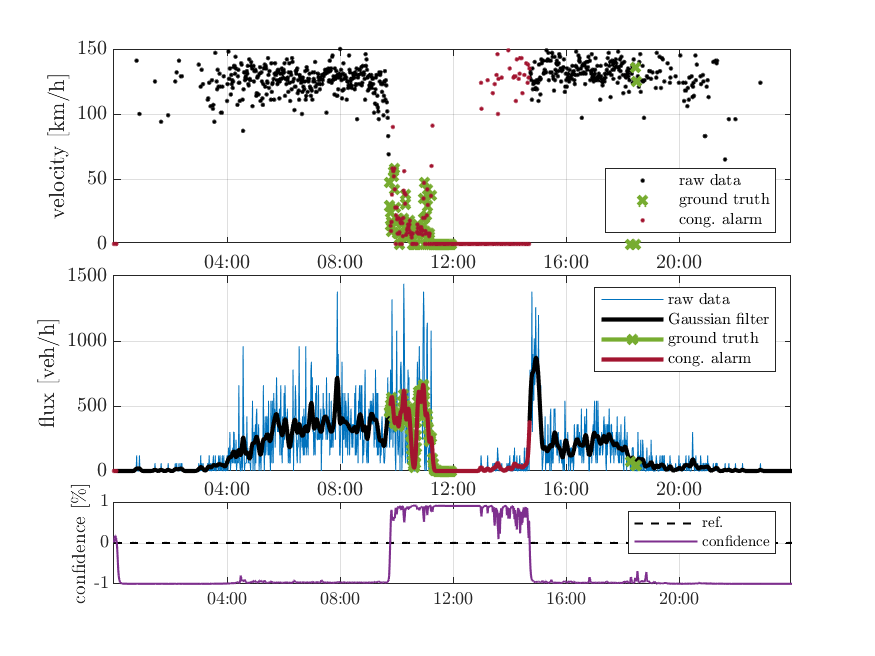}
        \includegraphics[width=\widthscale\linewidth]{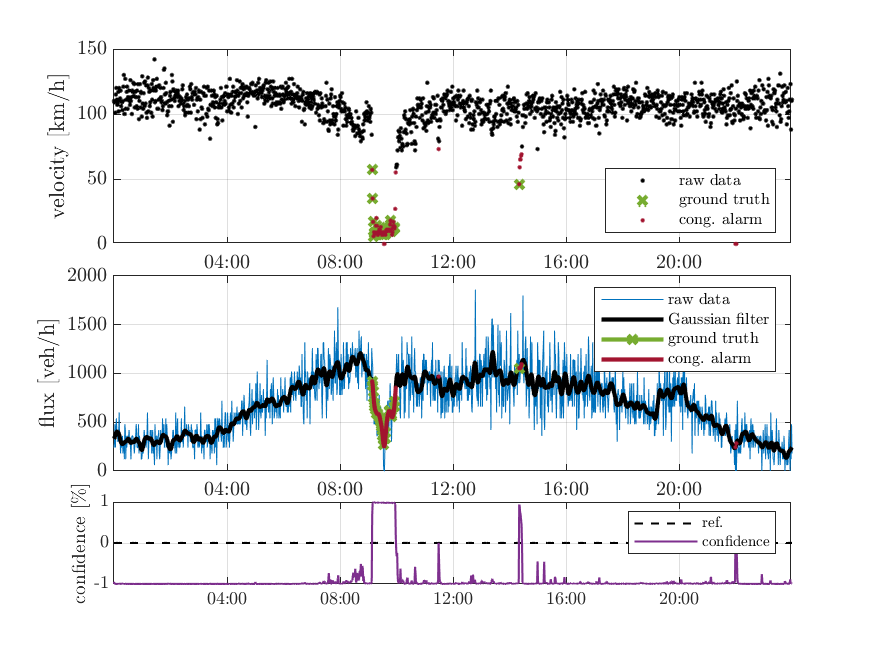}
        \caption{Day-length classification of congestion events. Each plot reports the velocity (top), the flux (middle) and the confidence of the model for the classification (bottom). Left: LF sensors; Right: HF sensors.}
        \label{fig:flunas-examples}
    \end{figure}
}

Figure~\ref{fig:flunas-examples} shows a few examples of day-length classification of congestion events for both LF and HF sensors. We report the velocity $v$, flux $f$, and the confidence of the classification (given as $\hat{o}$ rescaled in the interval $[-1, +1]$, \ie $\hat{o}\cdot 2-1$).
We can notice that when the drop is significant only in $v$ but not in $f$, the model shows a little uncertainty to classify the event (Figure~\ref{fig:flunas-examples}(top-left)); however, if the drops involve also $f$, then the model is able to classify the congestion even if they are short in time (Figure~\ref{fig:flunas-examples}(top-right)).
In particular circumstances, the model is able to classify critical events also when the ground truth is uncertain, like when the flux is close to zero and velocity is high (Figure~\ref{fig:flunas-examples}(bottom-left)). Note that this is a typical situation at night, with no congestion; finally, the model also keeps the congestion alarm on when the critical situation is vanishing: this comes from the fact that the model does not know the future, conversely to the regularization applied by the heuristics (Figure~\ref{fig:flunas-examples}(bottom-right)).

\subsection[Performance evaluation of the short-term prediction classifier]{Performance evaluation of the short-term prediction classifier $\FFFF_p$}
In order to give an idea about the actual possibility to predict congestion events, in Figure~\ref{fig:preflunas-examples} we shows four cases we have commonly observed.
Figure~\ref{fig:preflunas-examples}(top-left) shows a very favorable case in which we are able to predict the congestion 4 minutes before the actual formation. This is possible because the velocity starts lowering a bit before dropping down, together with the flux.
Figure~\ref{fig:preflunas-examples}(top-right) shows a case where a long-standing congestion is preceded by a confusing scenario in which the congestion alarm is on and off. In this case, the congestion pre-alarm is constantly on.
Figure~\ref{fig:preflunas-examples}(bottom-left) shows a case where prediction is quite hard, even for a trained human: congestion begins abruptly with a velocity drop, and the ANN is able to predict it only 1 minute in advance.
Finally, Figure~\ref{fig:preflunas-examples}(bottom-right) shows an uncertain situation which can likely evolve into a queue (but it does not, at least for some minutes). The pre-alarm is on, and even if this is formally incorrect (the congestion pre-alarm is not followed by a congestion alarm) the uncertainty fully justifies the ANN behavior.
{
    \def\widthscale{.49}
    \begin{figure}[h!]
        \centering
        \includegraphics[width=\widthscale\linewidth]{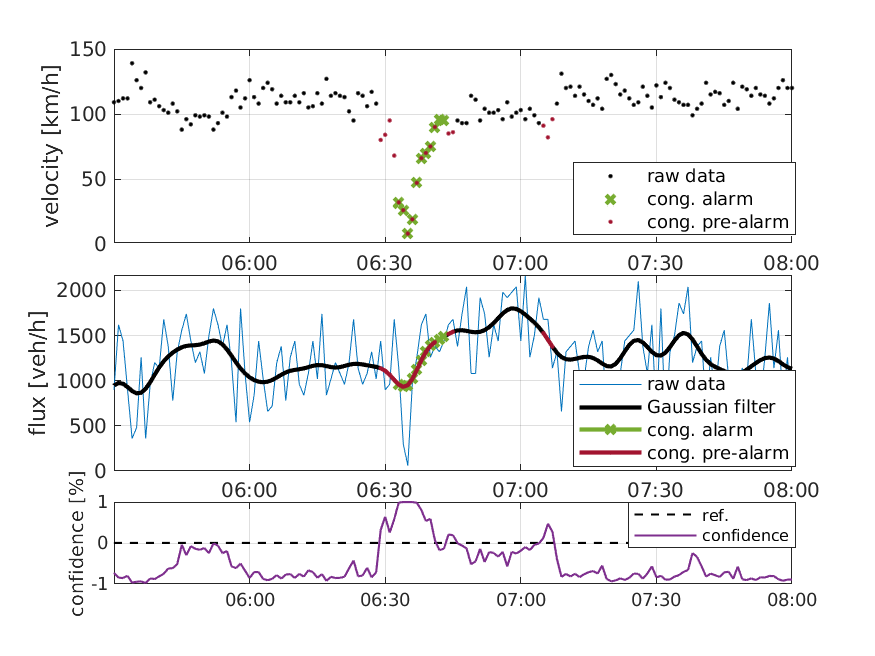}
        \includegraphics[width=\widthscale\linewidth]{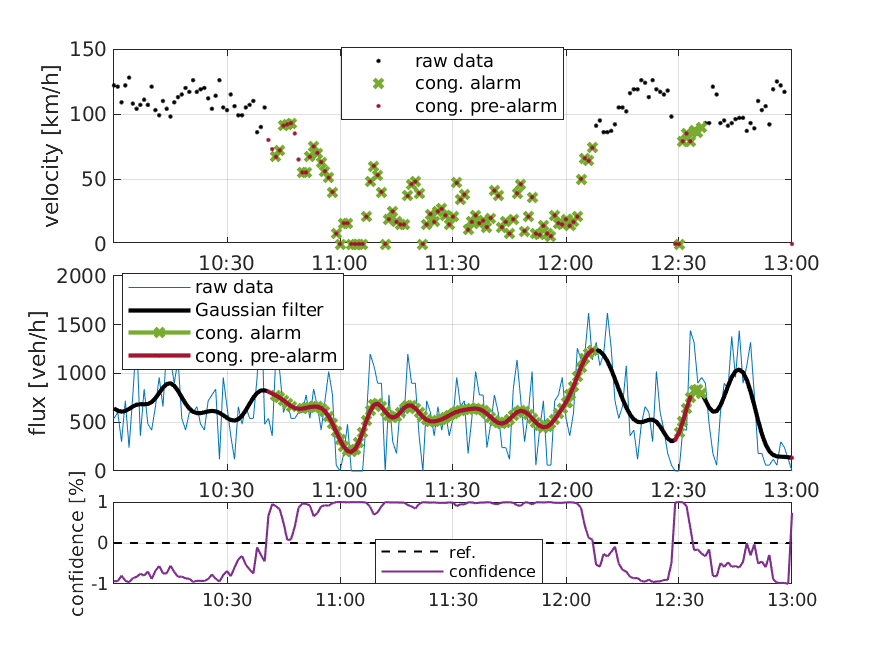}\\
        \includegraphics[width=\widthscale\linewidth]{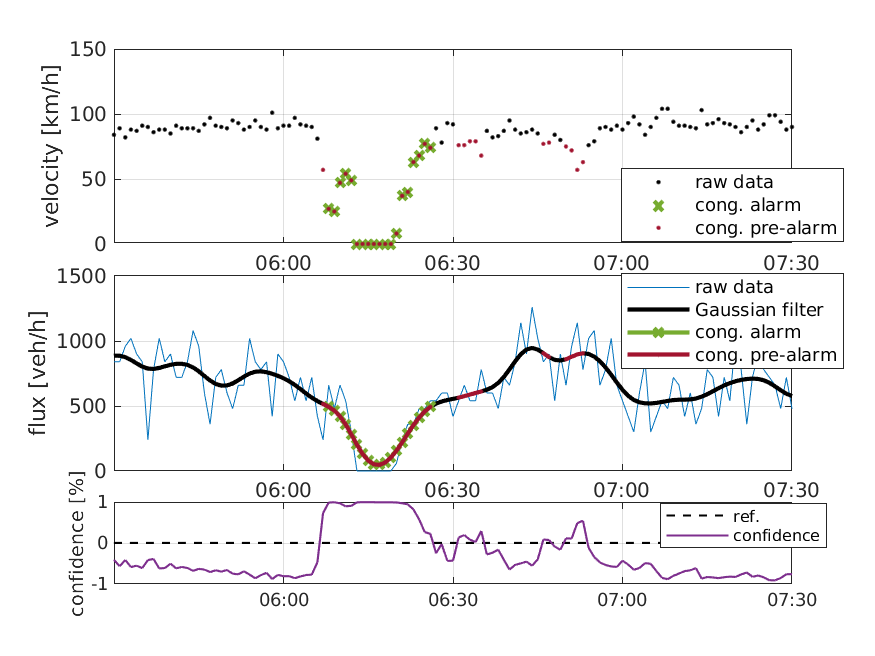}
        \includegraphics[width=\widthscale\linewidth]{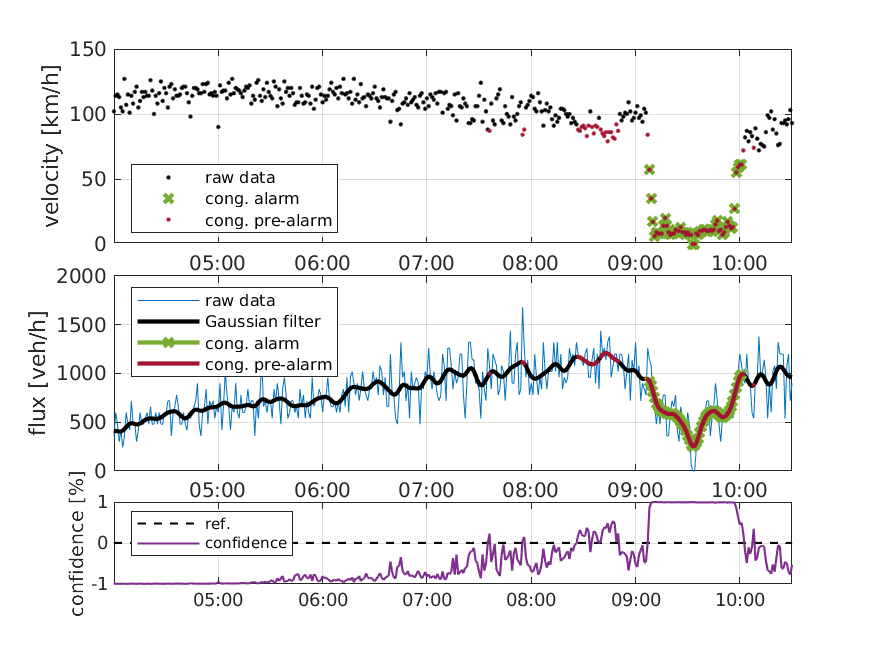}
        \caption{Top-left: at 6:30 a congestion event is correctly pre-alarmed four minutes in advance. Top-right: at 10:45 a congestion event begins intermittently while a solid pre-alarm shows up. Bottom-left: a difficult-to-predict congestion appears at 6:06. Bottom-right: at 8:30 an uncertain situation shows up, characterized by below-average velocity and normal flux, then after 9:00 a clear congestion is formed (pre-alarmed 1 minute in advance).}
        \label{fig:preflunas-examples}
    \end{figure}
}


\section{Computation of expected traffic volume}\label{sec:preconbo}

In this section, we tackle the problem of performing mid-term prediction (30 min) of the expected traffic volume by building an ANN-based predictor $\PPPP$.
Let us recall that, unlike in the previous section, here we aggregate sensors belonging to the same group (following the definition given in Section \ref{sec:discuss-data}), while working on the two distinct, but coupled, classes of vehicles (light and heavy).
This is important because the dynamics of the two classes are very different one from the other and, at the same time, they are strongly interconnected.

\subsection{Building the datasets}
Ground-truth data $\vec{y}$ can be easily produced for each (group of) sensor, only needing the real flux of light vehicles $\myvec{f}{\leg}$ and that of heavy vehicles $\myvec{f}{\pes}$ for the $\ttt$ upcoming minutes.
These compose the feature vector $\vec{x}$ as $(\myvec{f}{\leg}, \myvec{f}{\pes})$ and the corresponding ground truth $\vec y$ as:
\begin{equation}
    y_t= \left(  \frac1\ttt\sum_{s = t+1}^{t+\ttt}\myveccomp{f}{\leg}{s}, \  \frac1\ttt\sum_{s = t+1}^{t+\ttt}\myveccomp{f}{\pes}{s}\right).
\end{equation}

For what concerns the dataset creation, we selected the data collected during the entire 2021, corresponding to 348 day-wise readings (17 days are missing due to reported reading problems).
Due to the high number of available data, we decided to tune a predictor per group of sensors.
As for the previous training, we organized the features in day-length sequences, and we built a dataset $D^{(g)} = \{\myvecp{x}{d}; \myvecp{y}{d}\}$, $1 \leq d \leq 348$, for each group $g$.
We extracted roughly one month of samples from the dataset to form the test set while we used the remaining part as the training set.

\subsection{Training the model}

We trained a different model $\PPPP^{(g)}$ per group of sensors over its corresponding dataset $D^{(g)}$.
In particular, we focus on the four groups representing the inflow boundaries of the highway network under consideration: Venice ($g_1$) and Trieste ($g_2$) for the A4, Conegliano ($g_3$) for the A28, and Udine ($g_4$) for the A23, see Figure~\ref{fig:cartina-autovie}. 
$g_1$ is the only group deployed on a three-lane section.

By adopting the same nomenclature as from Figure~\ref{fig:methodology}, the choice of the input and output features requires $\nin = 2$ and $\npred = 2$, while $\nhid$ is a free parameter. 
Hence, we performed multiple training sessions to estimate the suitable size for the memory $\nhid$ of the LSTM.
Each training session was carried out with the ADAM optimizer, working with a variable learning rate, piece-wise decreasing over $5$ progressive learning eras of $50$ epochs each (for a total of $250$ epochs).

\begin{figure}[h!]
    \centering
    \includegraphics[width=0.5\linewidth]{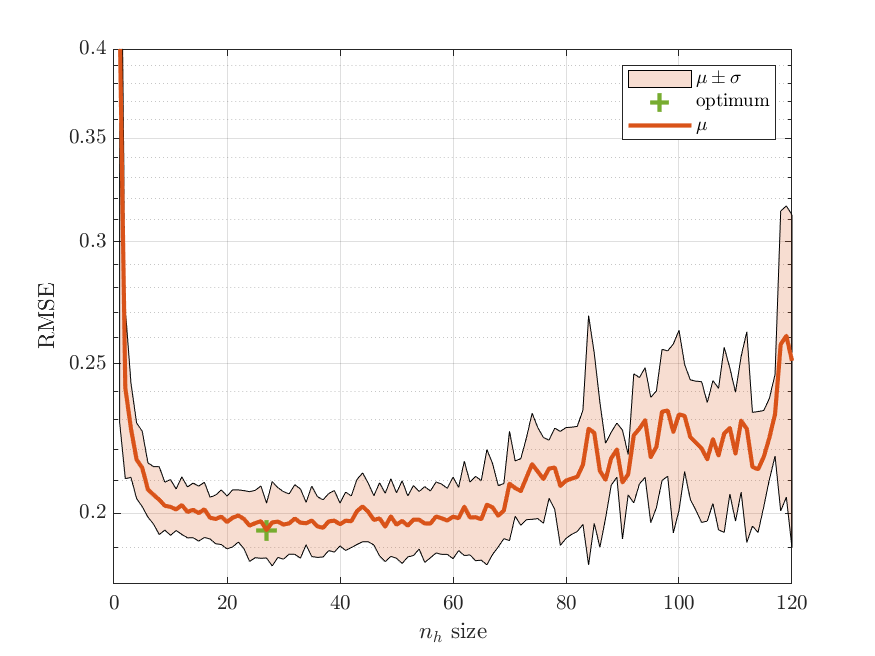}
    \caption{Value of the RMSE, corresponding to the training of $\PPPP$, as a function of the size $\nhid$.}
    \label{fig:RMSE-plot}
\end{figure}

Figure~\ref{fig:RMSE-plot} shows the value of the RMSE as a function of the free parameter $\nhid$.
We found $\nhid = 27$ being the best parameter to correctly capture the trend of the average flux; the average resulting error and the corresponding statistics are reported in Table~\ref{tab:preconbo-statitics}.
Note that the RMSE reported in Figure~\ref{fig:RMSE-plot} is evaluated on the normalized dataset, hence it is not comparable with results from Table~\ref{tab:preconbo-statitics}.
\begin{table}[h!]
    \centering
    \begin{tabular}{l||c|c|c|c}
        & \multicolumn{2}{c|}{Light vehicles} & \multicolumn{2}{c}{Heavy vehicles} \\
        & 2-lane & 3-lane & 2-lane & 3-lane \\
        \hline\hline
        Mean average-error & 42.74 & 76.60 & 11.74 & 25.06 \\
        Max average-error & 70.80 & 95.06 & 19.48 & 33.44 \\
        Mean standard-deviation & 41.72 & 62.96 & 12.04 & 23.94\\
        Mean max-error & 422.64 & 744.40 & 164.46 & 251.24 \\
        \hline
        Mean average-flux & 387.74 & 768.16 & 96.52 & 281.42 \\
        Mean max-flux & 5320.64 & 5864.48 & 1825.30 & 2047.10 \\
    \end{tabular}
    \caption{Error statistics in [vehicles/h] achieved by multiple trainings (with $\nhid = 27$) over the four datasets $D^{(g_i)}$ (average is performed on all minutes available in the dataset). Values are sorted by vehicle class and number of lanes. Average and maximum flux are also reported as benchmark values for  relative error statistics.}

   \label{tab:preconbo-statitics}
\end{table}

The question arises if it is really needed to train a different LSTM per group of sensors or if one ANN can serve all. To address this question we report in Figure~\ref{fig:preconbo-comparison} the error made by $\PPPP^{(g_i)}$ tested against datasets $D^{(g_j)}$, with $i,j = 1, \dots, 4$.
{
    \def\widthscale{.49}
    \begin{figure}[h!]
        \centering
        \includegraphics[width=\widthscale\linewidth]{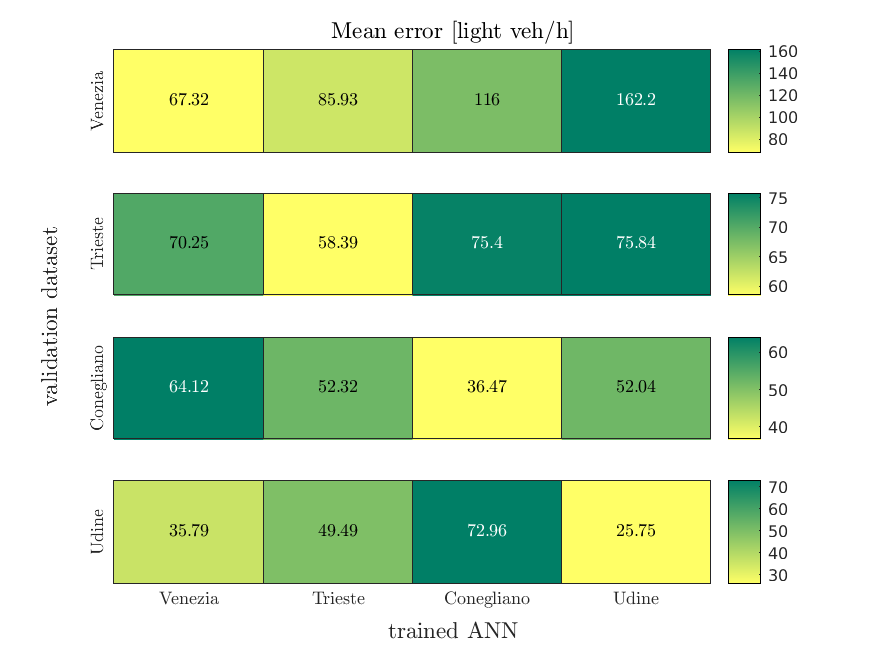}
        \hfill
        \includegraphics[width=\widthscale\linewidth]{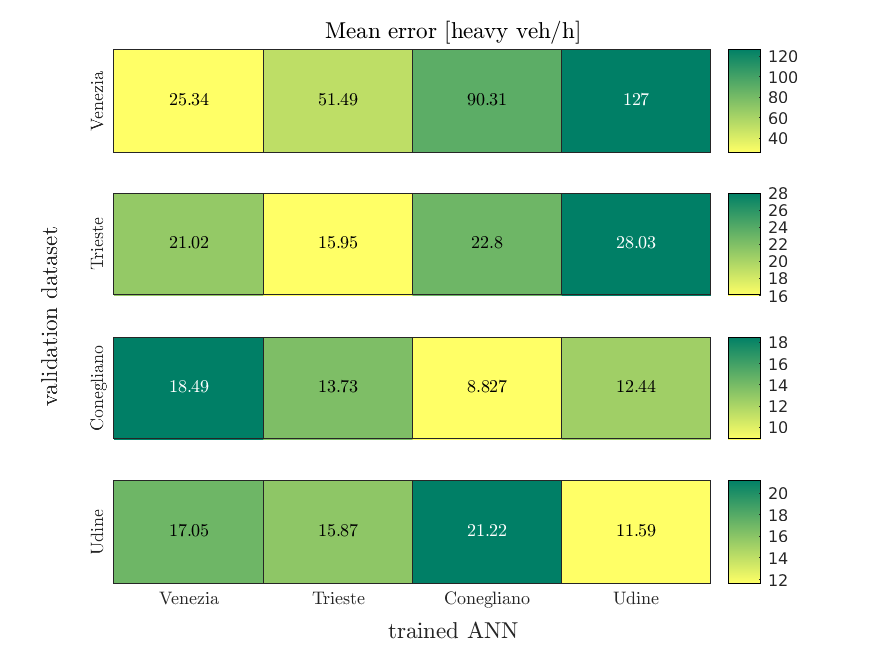}
        \caption{Mean error ($|y_t-o_t|$) in [vehicles/h] (light vehicles on the left, heavy vehicles on the right) obtained by applying the network $\PPPP^{(g_i)}, i \in \{1, 2, 3, 4\}$ ($x$-axis) on the mutual test-sets $D^{(g_j)}, j \in \{1, 2, 3, 4\}$ ($y$-axis). As it can be seen, maximum-by-row is achieved on the main diagonal, meaning that applying a specific ANN on a different dataset is not convenient. It is worth noticing that maximum-by-row does not necessarily correspond to the maximum-by-column (see, e.g., $i = 1, j = 4$); this phenomenon is due to the high regularity of some datasets.}
        \label{fig:preconbo-comparison}
    \end{figure}
}
We see that it is highly suggested to train as many ANN as groups of sensors are considered.

\subsection{Performance evaluation}
We consider the group of sensors $g_2$ as an example for the performance evaluation.
Figure~\ref{fig:preconbo-trieste-histos} shows the histograms of the errors (in [vehicles/h]) made running the prediction every minute available in the test set.
{
    \def\widthscale{.49}
    \begin{figure}[h!]
        \centering
        \includegraphics[width=\widthscale\linewidth]{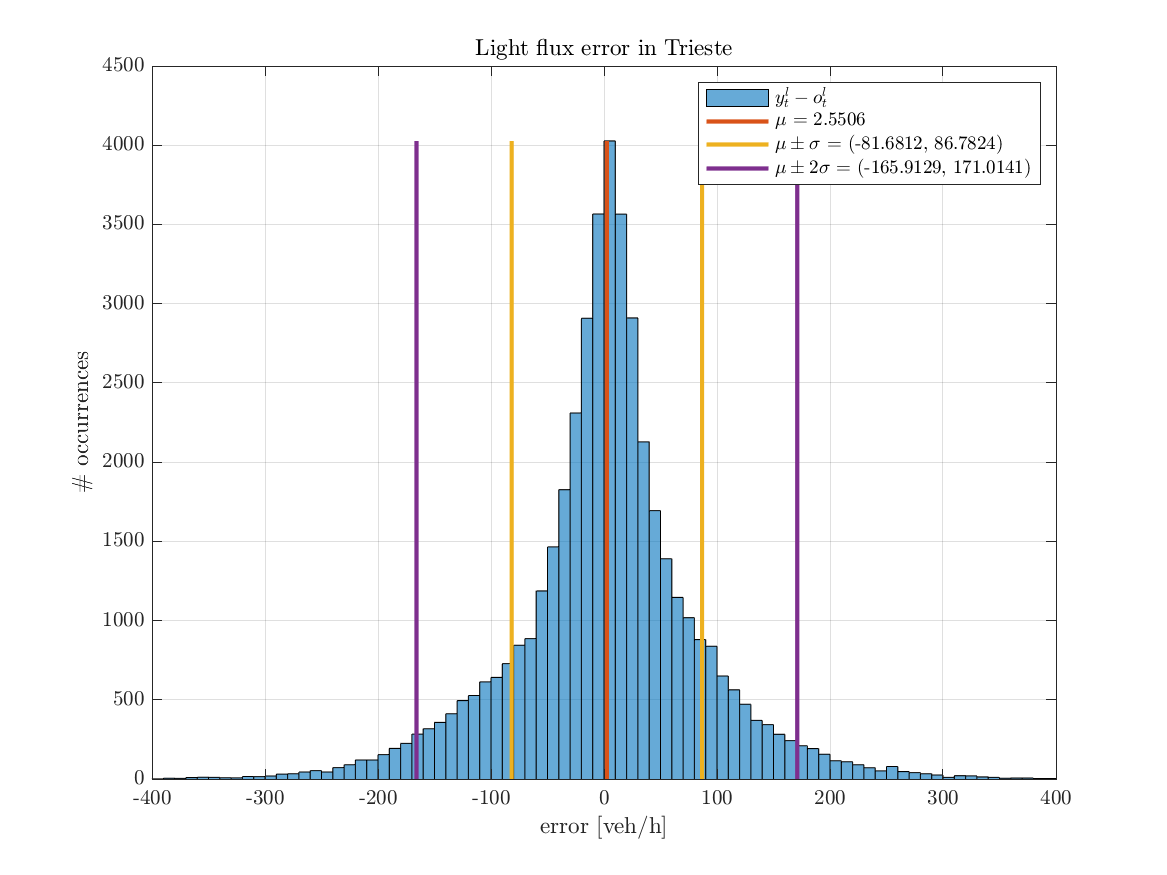}
        \hfill
        \includegraphics[width=\widthscale\linewidth]{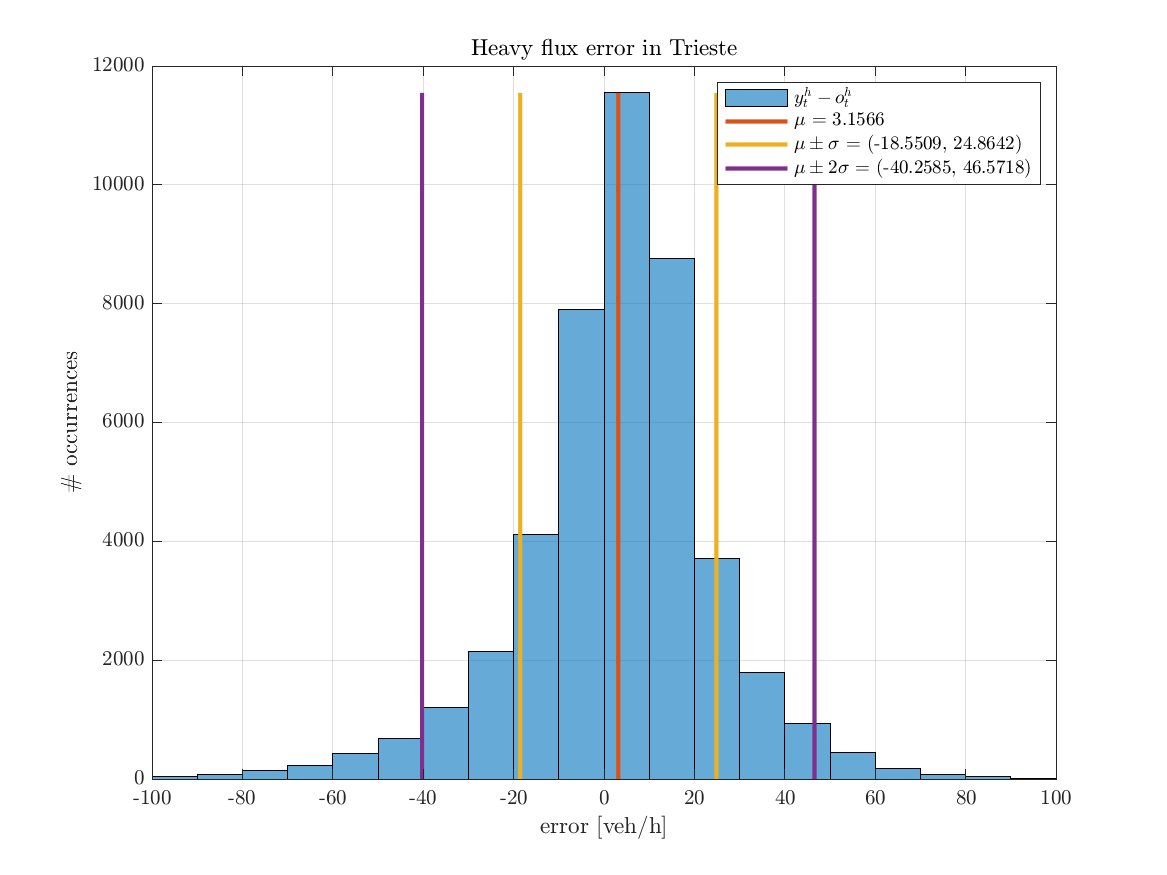}
        \caption{Error histogram of the predictions of $\PPPP^{g_2}$ over its test set. Light (left) and heavy (right) vehicles. Error is almost symmetrical w.r.t.\ 0.}
        \label{fig:preconbo-trieste-histos}
    \end{figure}
}
Figure~\ref{fig:preconbo-trieste-samples} shows instead the result of $\PPPP^{g_2}$ on an entire day.
Every minute, the ground truth (i.e.\ the total number of vehicles passed in the next 30 minutes, in [vehicle/h]) is compared with its ANN-predicted value, 
and the offset is evaluated. 
Real flux $\vec{f}$ is reported along with its \emph{a posteriori} regularization for reference.
It can be seen again that the error in prediction is almost symmetrical and it usually corresponds to less than 200 vehicles/h for light vehicles and much less for heavy vehicles.
{
    \def\widthscale{.49}
    \begin{figure}[h!]
        \centering
        \includegraphics[width=\widthscale\linewidth]{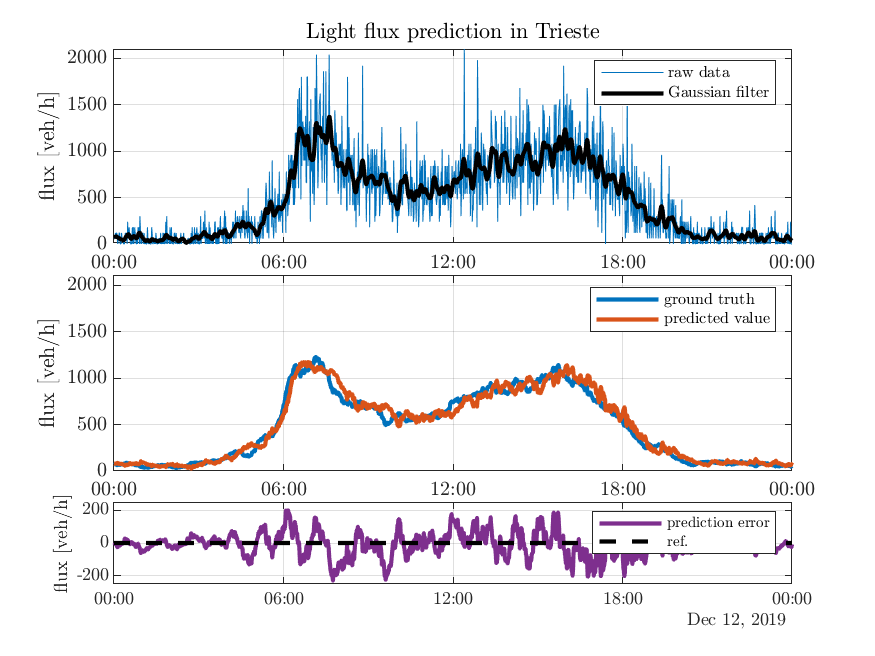}
        \hfill
        \includegraphics[width=\widthscale\linewidth]{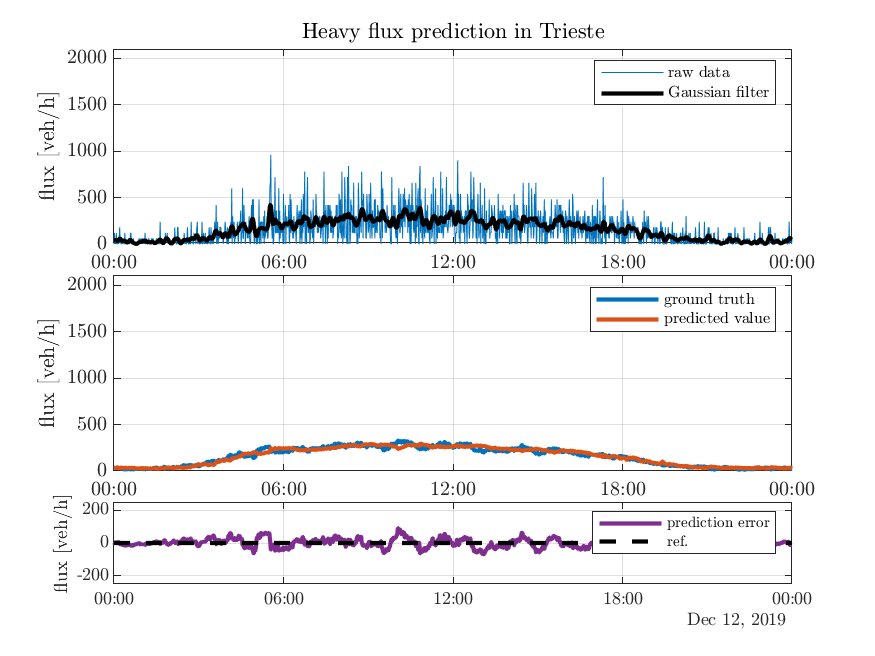}
        \caption{$\PPPP^{g_2}$-predicted vs.\ actually measured flux for 30 mins in the future, minute by minute, for light (left) and heavy (right) vehicles.}
        \label{fig:preconbo-trieste-samples}
    \end{figure}
}

\begin{figure}[h!]
        \centering
        \includegraphics[width=.9\linewidth]{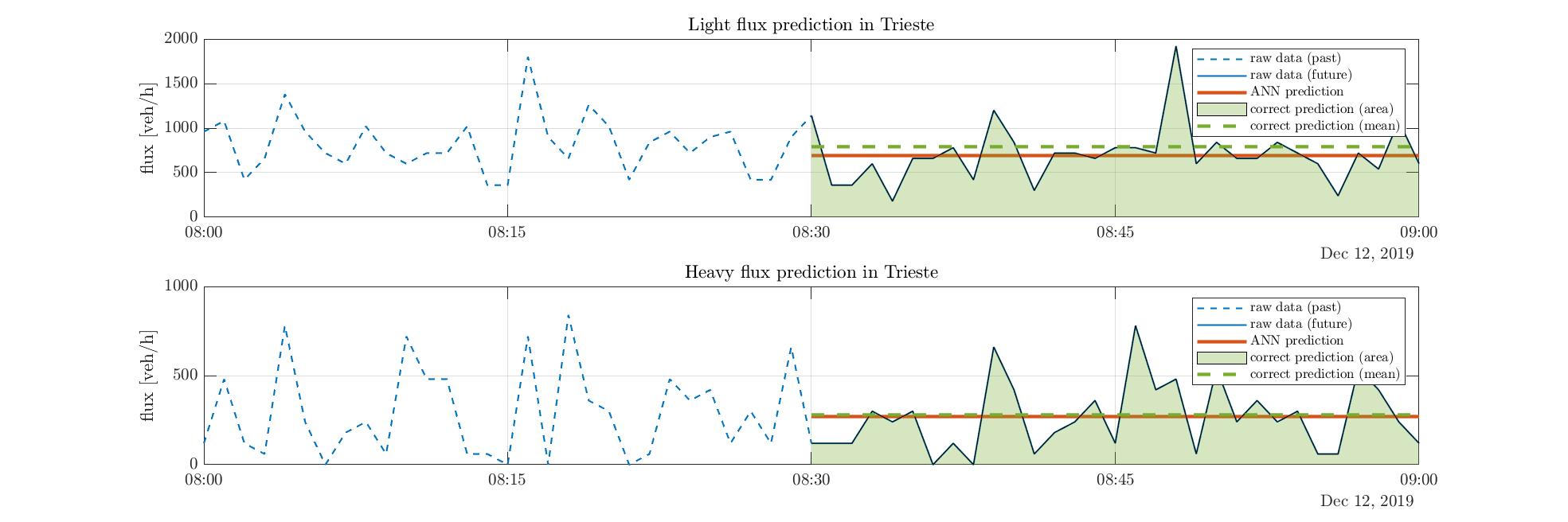}
        \caption{Zoom of a single-minute prediction of the data presented in Figure~\ref{fig:preconbo-trieste-samples}. Prediction is performed at time $t$=8:30, where dashed data are available. The predicted value is given by the area beneath the orange line while the target of the prediction is represented by the value of the green shaded area (represented also as the green dashed line for a better comparison).}
        \label{fig:preconbo-trieste-samples-zoom}
    \end{figure}

Finally, Figure~\ref{fig:preconbo-trieste-samples-zoom} shows an example of a single-minute prediction.


\section{Feeding traffic models with ML-enriched data}\label{sec:models}
The tools introduced in the previous sections are useful \textit{per se}, but they can also be used in combination with other, more classical, tools for traffic estimation and forecast. 
In this section, we try to get advantage of the pieces of information extrapolated by data in order to improve the accuracy of the macroscopic differential models.

\subsection{Nowcast}

In this section we explore the possible advantages of inverting the fundamental diagram when estimating the current status of traffic density by means of a macroscopic differential model. 
As mentioned in Section \ref{sec:introduction}, we split the whole road $[\xmin,\xmax]$ into consecutive segments delimited by fixed sensors. 
We aim at estimating the traffic density at the current time $t_0$, therefore we start the simulation at a previous time $t_0-\Delta \tpast$, with $\Delta \tpast>0$, assuming an empty road at that time. 
Let us denote by $N_S$ the number of road segments (sensors are $N_{S}+1$).
Each road segment $S^k:=(s^k,s^{k+1})$, $k=1,\ldots,N_S$, is delimited by two fixed sensors, located at $x=s^k$ and $x=s^{k+1}$, respectively. 

We begin with a simplified setting, then we move to the real one.

\subsubsection{Simplified setting}
We consider the classical single-lane single-class LWR model on each road segment $S_k$ 
\begin{equation}\label{LWR}
\left\{
\begin{array}{ll}
\partial_t\rho^k(x,t) + \partial_x f(\rho^k(x,t))=0, & x\in S^k, \quad t\in (t_0-\Delta \tpast,t_0) \\
\rho^k(x,t_0)\equiv 0, & x\in S^k \\
\rho^k(s_k,t)=\rhoin^k(t), & t\in [t_0-\Delta \tpast,t_0] \\
\rho^k(s_{k+1},t)=\rhoout^k(t), & t\in [t_0-\Delta \tpast,t_0]
\end{array}
\right.
\end{equation}
where $\rho^k\in[0,\rho_{\textsc{max}}]$ is the vehicle density for some maximal density $\rho_{\textsc{max}}^k>0$, and $\rho\mapsto f(\rho)$ is the fundamental diagram. 
Let us assume, as usual, that $\rho\mapsto f(\rho)$ is concave and denote by $\sigma$ the argmax of $f$, i.e.\ $f(\sigma)=\max_\rho f(\rho)$ (see Figure~\ref{fig:FDgeneric}).
Equation \eqref{LWR} is defined independently on each segment.
Let us also recall that the relation \eqref{f=rhov} holds true.

Equation \eqref{LWR} is usually solved by numerical approximation. Let us introduce a grid in the domain $S^k\times [t_0-\Delta \tpast,t_0]$, with space step $\Delta x$ and time step $\Delta t$.
The time interval is divided into $N_t$ intervals, while each segment $S^k$ is divided into $N_x^k$ cells of length $\Delta x$ and the approximate average density in cell $C^k_j$, $j=1,\ldots,N_x^k$ at time $n$ is denoted by $\rho^{k,n}_j$.
Any conservative numerical scheme \cite{levequebook} for \eqref{LWR} has the form
\begin{equation}\label{conservativescheme}
    \rho^{k,n+1}_j=\rho^{k,n}_{j}-\frac{\Delta t}{\Delta x}\Big(F(\rho^{k,n}_{j},\rho^{k,n}_{j+1})-F(\rho^{k,n}_{j-1},\rho^{k,n}_{j})\Big),\qquad j=1,\ldots,N_x^k,
\end{equation}
where $F$ is the \emph{numerical flux} (i.e.\ an approximation of the flux $f$ at the interface between two consecutive cells). 
For example, in the case of the Godunov scheme, we have
\begin{equation}\label{GodunovFlux}
F(\rho_-,\rho_+):=
\left\{
\begin{array}{ll}
\min\{f(\rho_-),f(\rho_+)\} & \textrm{if } \rho_-\leq \rho_+ \\
f(\rho_-) & \textrm{if } \rho_->\rho_+ \textrm{ and } \rho_-<\sigma \\
f(\sigma) & \textrm{if } \rho_->\rho_+ \textrm{ and } \rho_- \geq \sigma \geq \rho_+ \\
f(\rho_+) & \textrm{if } \rho_->\rho_+ \textrm{ and } \rho_+>\sigma
\end{array}
\right. .
\end{equation}

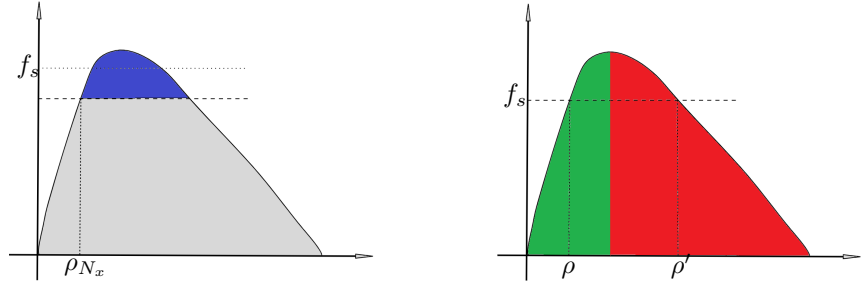
\begin{figure}[h!]
 \centering
 \begin{overpic}[width=.4\linewidth]{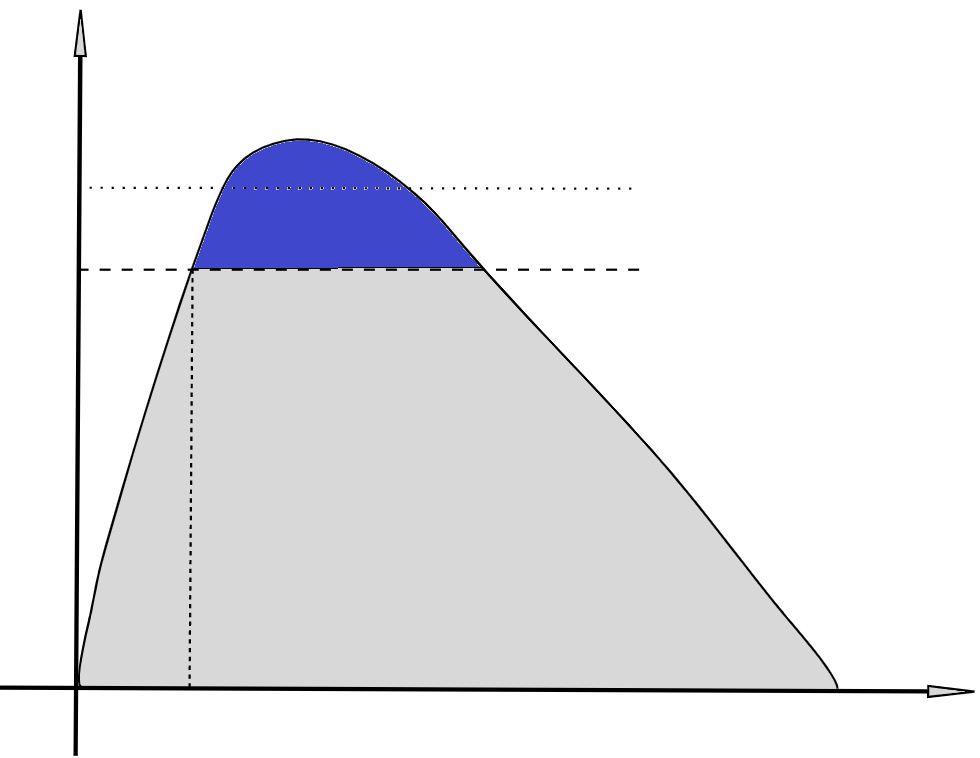} 
 \put(15,3){$\rho_{N_x}$} 
 \put(2,57){$f_{s}$} 
 \end{overpic} \hskip1.5cm
 \begin{overpic}[width=.4\linewidth]{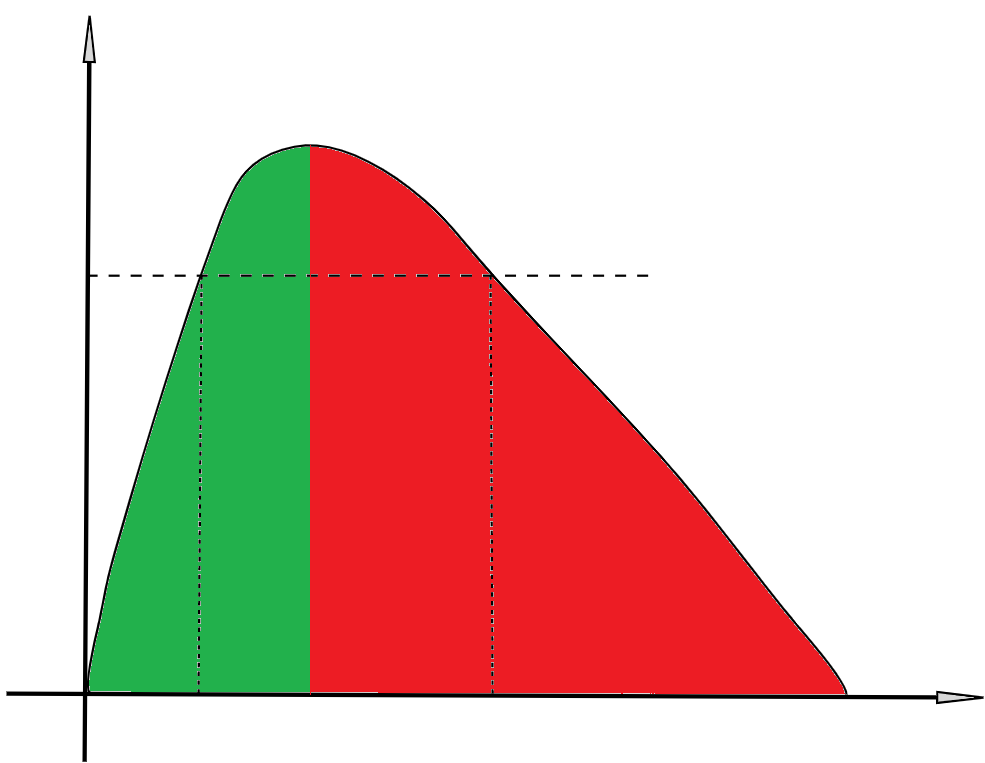} 
 \put(18,2){$\rho$} \put(48,2){$\rho'$} 
 \put(2,49){$f_s$} 
 \end{overpic} 
\caption{Fundamental diagram $f=f(\rho)$. (Left) Example of nonadmissible sensor flux datum for the numerical flux \ref{GodunovFlux}: no flux $f_s>f(\rho_{N_x})$ is compatible with any boundary condition $\rhoout$. (Right) Given the sensor flux datum $f_s$, the corresponding density $\rho$ or $\rho'$ is chosen depending on the presence or not of the congestion.}
 \label{fig:tau_di_rho}
\end{figure}

Let us consider, e.g., the right boundary condition of a given segment $S^k$, which corresponds to the left boundary condition of the following segment $S^{k+1}$. 
Let us also drop the index $k$ from $\rho$ for readability: 
if $j=N_x^k$, one could follow a
\begin{itemize}
\item[-] \emph{flux-based approach}: directly inject in the scheme the flux datum $f_{s^k}$ measured by the sensor across the interface $S^k| S^{k+1}$ in place of the numerical outgoing flux $F(\rho^{n}_{N_x},\rhoout^n)$, without estimating the density $\rhoout$;
\item[-] \emph{density-based approach}: evaluate the numerical flux $F(\rho^{n}_{N_x},\rhoout^n)$, by estimating the density $\rhoout$. 
\end{itemize}
The two ways are in principle both correct and the first one seems to be more practical since sensors provide flux data only (i.e.\ density is not available at all).
On the other hand, flux data $f_s$ provided by the sensor are not always compatible with the solution carried by the numerical scheme. In fact, any flux $f_s$ outside the set of admissible values
\begin{equation}\label{admissiblefluxes}
\left\{F(\rho^{n}_{N_x},\rho):\rho\in[0,\rho_{\textsc{max}}]\right\}
\end{equation}
is not compatible with \eqref{LWR} and leads to negative densities, since more mass comes out than it is available in the road, see Figure~\ref{fig:tau_di_rho}-left.
The question arises how to enforce the compatibility of the sensor data: a natural solution is to project the sensor data into the admissible set of flux data \eqref{admissiblefluxes}.

Regarding the density-based approach, the idea is to use the algorithm introduced in Section \ref{sec:flunas} to help invert the concave fundamental diagram, i.e.\ passing from the flux data to the density data by duly distinguishing between free and congested regimes. 
More precisely, given the sensor flux datum $f_s< f_{\textsc{max}}$, the choice between $\rho$ and $\rho'$, with $\rho'\neq\rho$, $f(\rho)=f(\rho')=f_s$, is taken depending on the presence or not of the congestion, see Figure~\ref{fig:tau_di_rho}-right.

\medskip

In order to discuss the difference between the two approaches, we devise a simple numerical test. We consider an infinitely-long single-lane road. As it is usually done in the mathematical literature, we normalize both density and velocity in the interval $[0,1]$ and we choose the fundamental diagram as $f(\rho)=\rho(1-\rho)$. 
In this case, the maximal flux is 0.25 and it is achieved for $\rho=\sigma:=0.5$. 
Two sensors are located at $x=s^k:=0.45$ and $x=s^{k+1}:=0.8$. 

At initial time $t=t_0-\Delta \tpast$ the road has constant density $\rho=0.45$ for $x<0.52$ while the rest of the road is empty ($\rho=0$).
Immediately after the initial time, an accident occurs at $x=b:=0.6$ and a bottleneck is formed between the two sensors.  

\begin{figure}[h!]
 \centering
 \begin{overpic}[width=.49\linewidth]{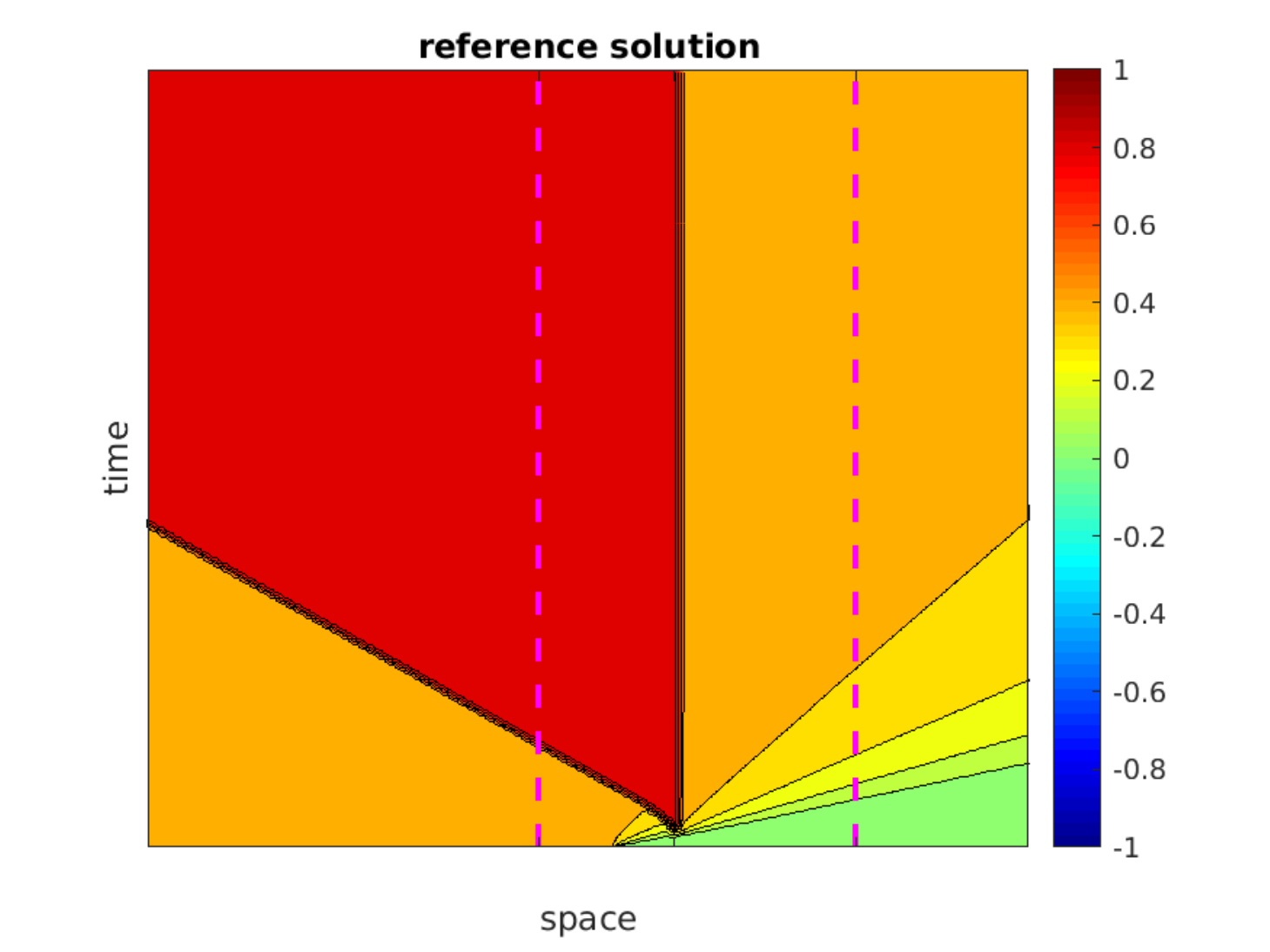}
 \put(40,4){$s^k$} \put(64,4){$s^{k+1}$} \put(52,4){$b$} 
 \end{overpic} 
\caption{Academic test. Reference solution}
 \label{fig:test_flunas_academic_reference_solution}
\end{figure}

Figure~\ref{fig:test_flunas_academic_reference_solution} shows the reference simulation, i.e.\ what we assume it is really happening on the road and that we would ideally like to reproduce with data at our disposal: at the initial time, a rarefaction fan immediately appears at $x=0.52$ and the right part of the road starts populating. 
When enough vehicles have reached the bottleneck at $x=b$, a queue appears and starts back-propagating. The queue reaches the sensor at $s^k$ and continues back-propagating. From the bottleneck on, vehicles set off at maximal flux ($\rho=\sigma$) and proceed normally.
 \begin{figure}[h!]
 \centering
  \includegraphics[width=.49\linewidth]{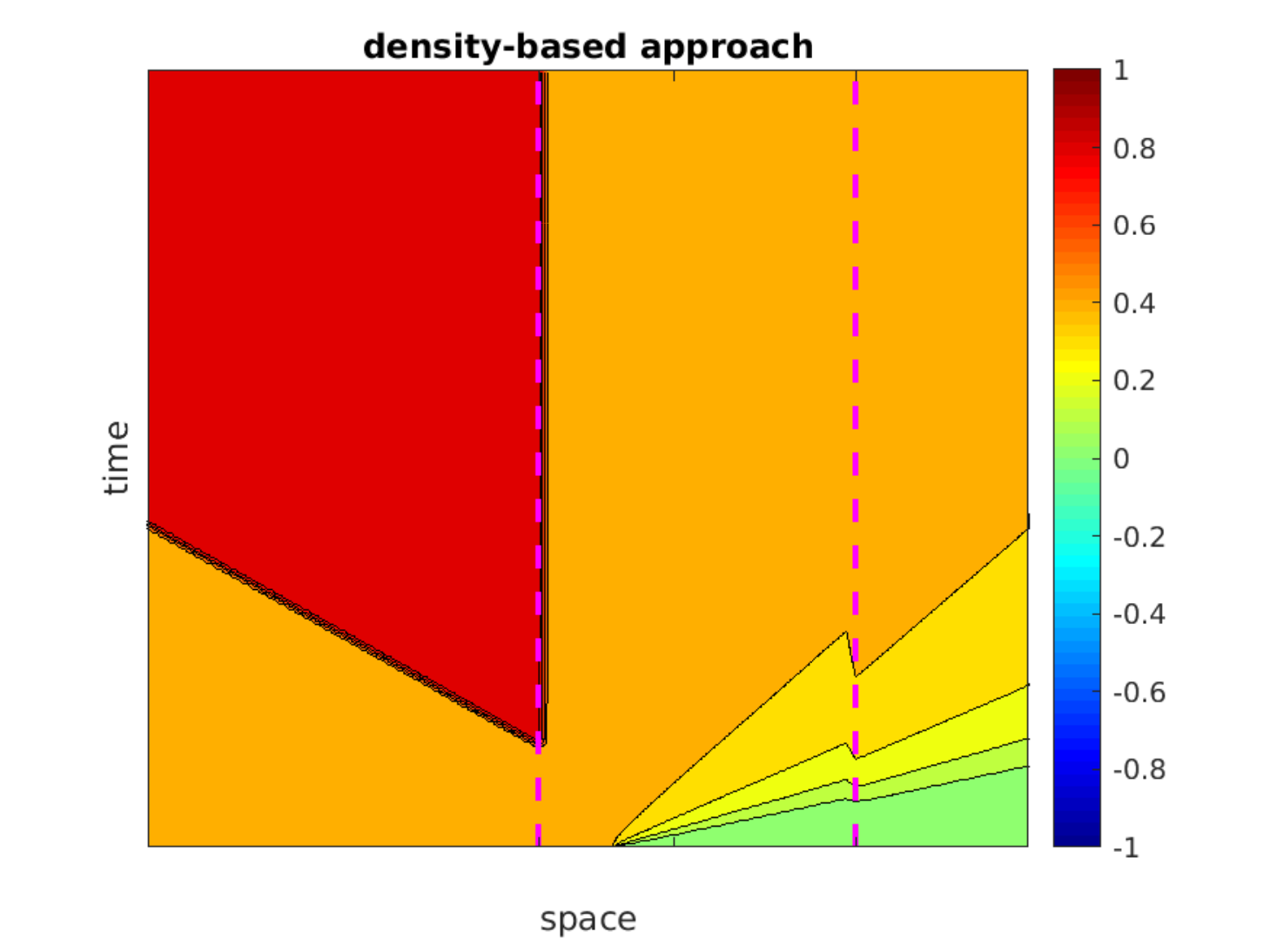}
  \includegraphics[width=.49\linewidth]{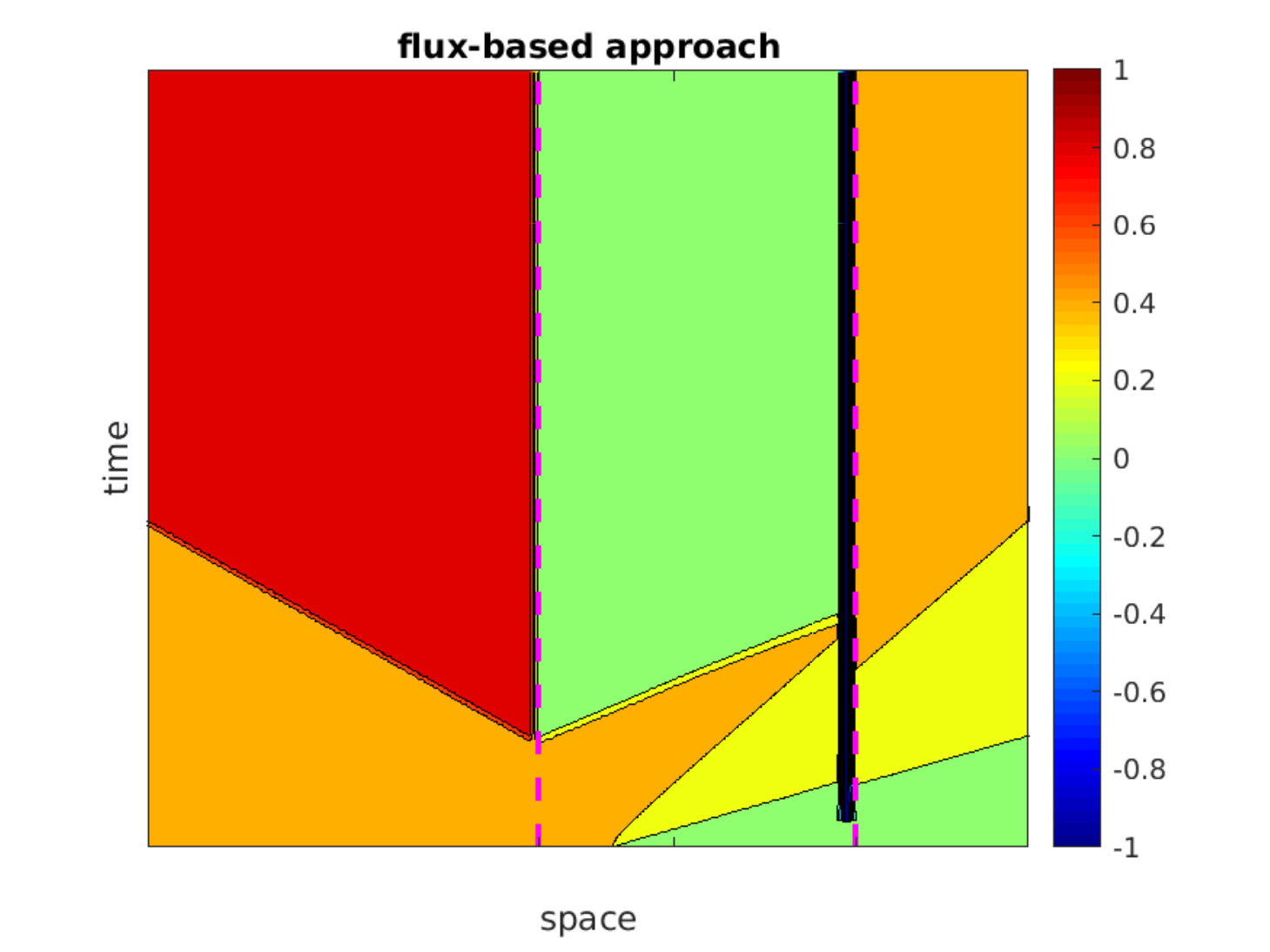}
   \caption{Academic test. Left: density-based approach. Right: flux-based approach}
 \label{fig:test_flunas_academic}
\end{figure}

It is plain that, if the traffic is observed only at sensors, we cannot be able to perceive the accident and the formation of the bottleneck in its actual position. The effects of the accident will be visible only when the queue reaches the sensor at $x=s^k$.

Figure~\ref{fig:test_flunas_academic}-left show the result obtained by the \emph{density-based} approach. 
When the queue is perceived at $x=s^k$, the recorded flux $f_{s^k}=0$ is correctly translated into the maximal density $\rho=1$ inverting the fundamental diagram. Therefore, the queue continues back-propagating while, for $x>s^k$, the traffic restarts with maximal flux. 
At $x=s^{k+1}$ whatever approach is used, the solution is the same and the traffic keeps going with the same dynamics.

Figure~\ref{fig:test_flunas_academic}-right show instead the result obtained by the \emph{flux-based} approach. 
As before, when the queue reaches $x=s^k$ the sensor registers a null flux there, and a queue starts back-propagating. Since the flux through the sensor is null, no one moves from the sensor on, and the space between the sensors becomes empty in a short time. At this point, the flux data at the right sensor becomes incompatible with the traffic condition (the road is empty but the sensor perceives moving vehicles). 
Therefore, a negative density appears at $x=s^{k+1}$. For $x>s^{k+1}$ traffic dynamics restart correctly with the measured flux.

In conclusion, in this case, the density-based approach is preferable since it catches with better precision the real scenario, and this is achieved since the density-based approach actually puts in the simulation additional information about the system other than the naked flux data, i.e.\ the discrimination free/congested scenario.

\subsubsection{Real setting}
In this section, we consider real sensor data provided by Autovie Venete, see Section \ref{sec:discuss-data}.
The major difference with respect to the previous test is that now we deal with a three-lane highway and two classes of vehicles (light and heavy), with coupled dynamics. The
LWR-like model is generalized to this case by a system of PDEs
\begin{equation}\label{eq:modelLH}
\left\{
  \begin{array}{ll}
    \partial_t\rhoL^k + \partial_x f_{\leg}(\rhoL^k,\rhoP^k) =0, & x\in S_k, \quad t\in (t_0-\Delta \tpast,t_0)  \\ [2mm]
    \partial_t \rhoP^k + \partial_x f_{\pes}(\rhoL^k,\rhoP^k) =0, &  x\in S_k, \quad t\in (t_0-\Delta \tpast,t_0) 
\end{array}
\right.
\end{equation}
where $\rhoL$, $f_{\leg}$ and $\rhoP$, $f_{\pes}$ are the density and flux of light and heavy vehicles, respectively. 
Equation \eqref{eq:modelLH} is complemented with initial and boundary conditions as in \eqref{LWR}.
Moreover, the two classes of vehicles do not share the road in the same manner, being heavy vehicles not allowed in the fastest lane.
The coupled dynamics, with uneven space occupancy, was already derived in  \cite{briani2021AXIOMS} and we refer the reader to that paper for both the mathematical and numerical details.
Here we just recall that we consider a \emph{phase transition} (cf.\ \cite{colombo2002SIAP, colombo2010JHDE, dellemonache2021AX}) due to the presence of two states of the system, see Figure~\ref{fig:2phases}:
\begin{itemize}
    \item The \emph{partial-coupling phase}, when heavy vehicles influence the dynamics of light ones but not vice versa.
    Light vehicles are then mainly in the fast lane and heavy vehicles are independent of them. 
    In this case, the two equations in the system \eqref{eq:modelLH} are partially coupled, i.e.\ $f_{\pes}$ only depends on $\rhoP$,
    \begin{equation*}\label{eq:model_PC}
	\begin{cases}
	\partial_t\rhoL + \partial_x f_\leg(\rhoL,\rhoP)=\, 0, 
	\smallskip\\
	\partial_t \rhoP + \partial_x f_\pes(\rhoP)=\, 0.
	\end{cases}
    \end{equation*}
    \item The \emph{full-coupling phase}, when light vehicles are too much to fit the fast lane only and then invade the slow lane(s), influencing the dynamics of heavy vehicles. 
    In this case, the two equations are fully coupled and fall in the general form of the system \eqref{eq:modelLH} . 
\end{itemize}
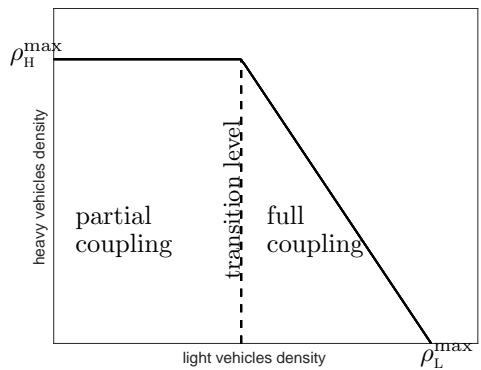
\begin{figure}[h!]
    \centering
    \begin{overpic}
	[width=0.6\columnwidth]{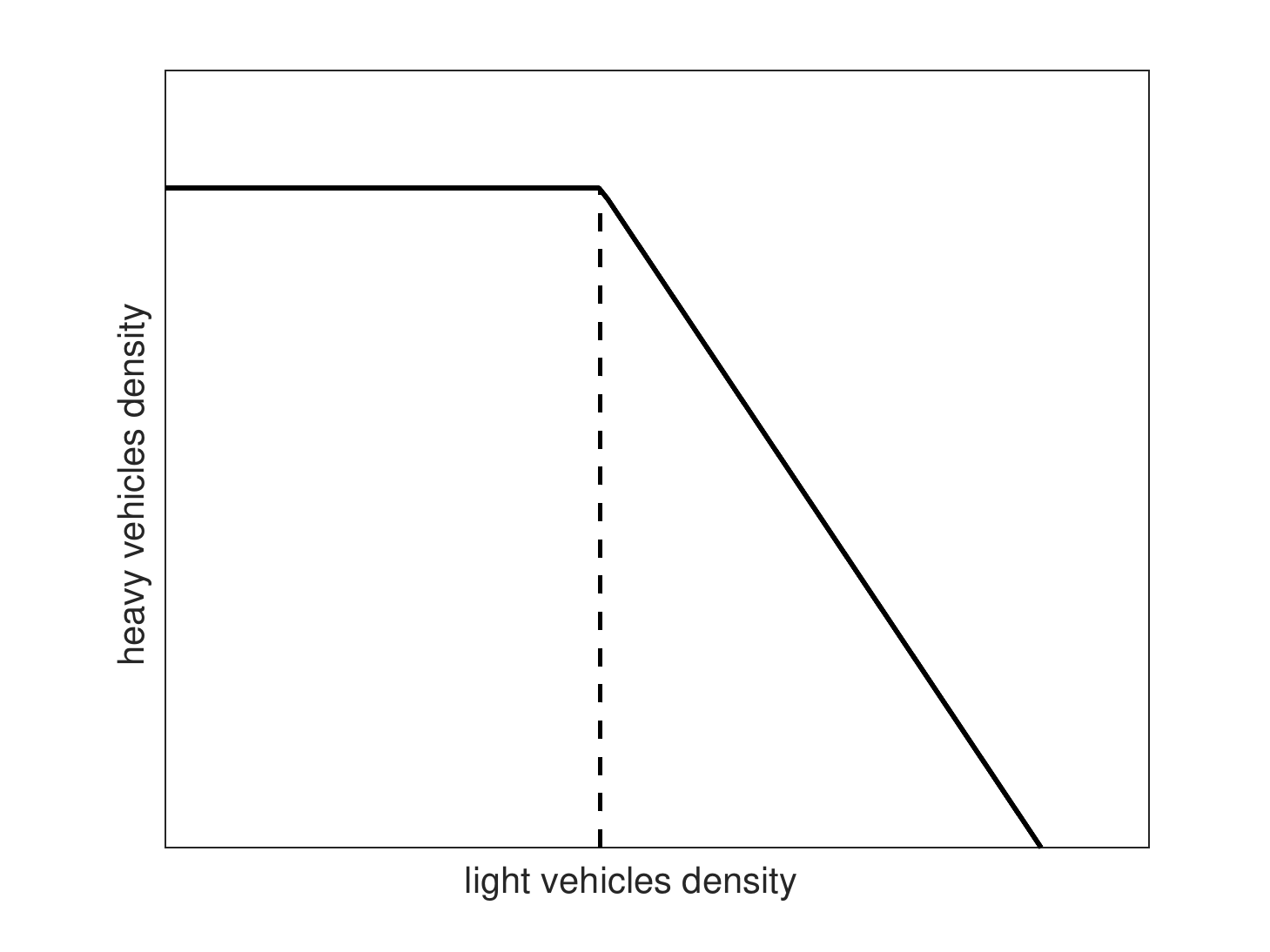}
	\put(17,30){partial}  \put(17,25){coupling}
	\put(52,30){full}  \put(52,25){coupling}
	\put(44,20){\rotatebox{90}{\small transition level}}
	\put(80,5){$\rhomax$}
	\put(5,60){$\mumax$}
    \end{overpic}
    \caption{The two phases of the model \eqref{eq:modelLH}: the density of light vehicles determines if the dynamics is partially or fully coupled.}
    \label{fig:2phases}
\end{figure}

\begin{figure}[h!]
\centering
			\begin{overpic}
			[width=.49\linewidth]{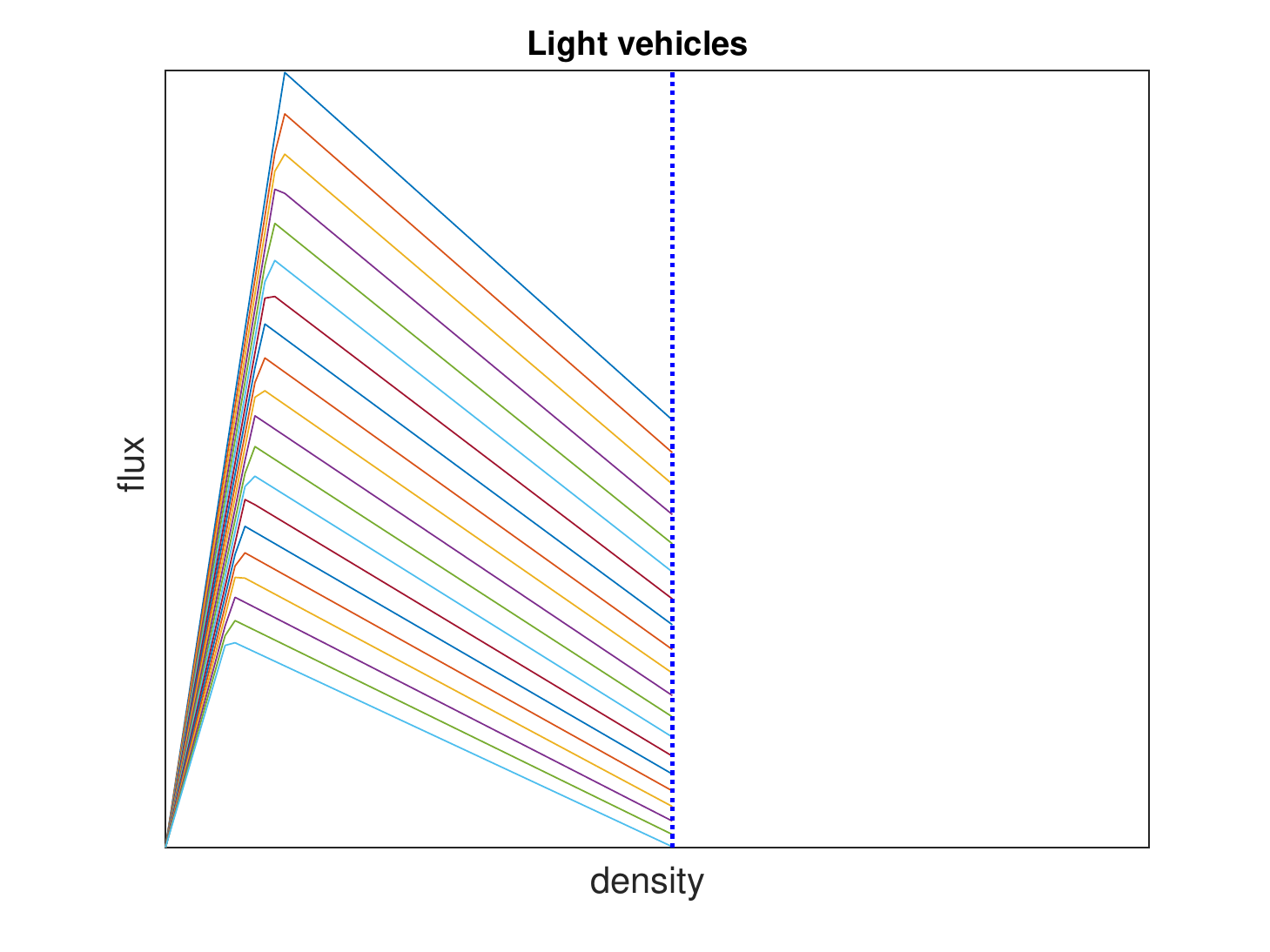}
			\put(50,47){\rotatebox{90}{\tiny transition level}}
			\end{overpic}
			\includegraphics[width=.49\linewidth]{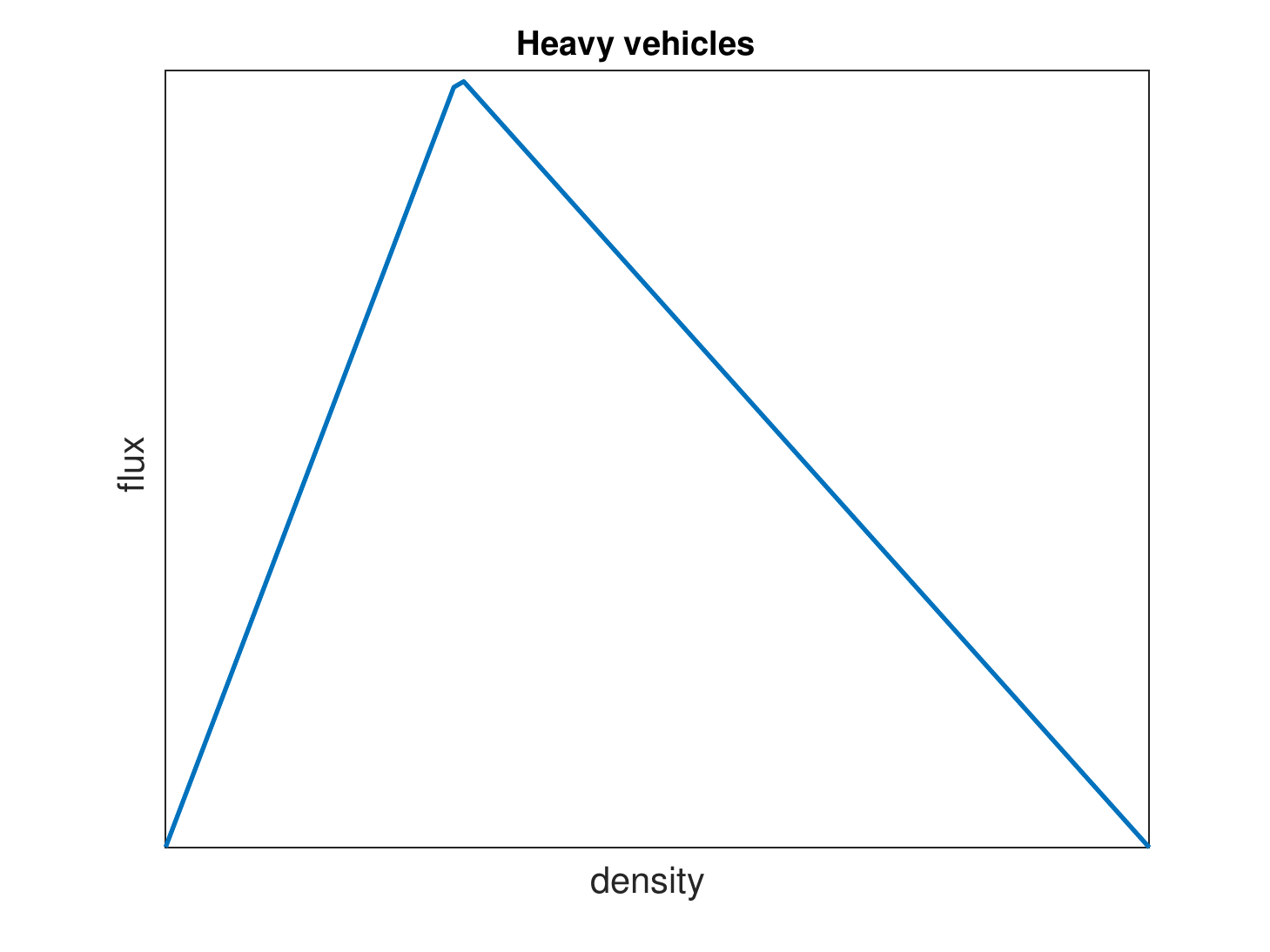}\\
			\begin{overpic}
				[width=.49\linewidth]{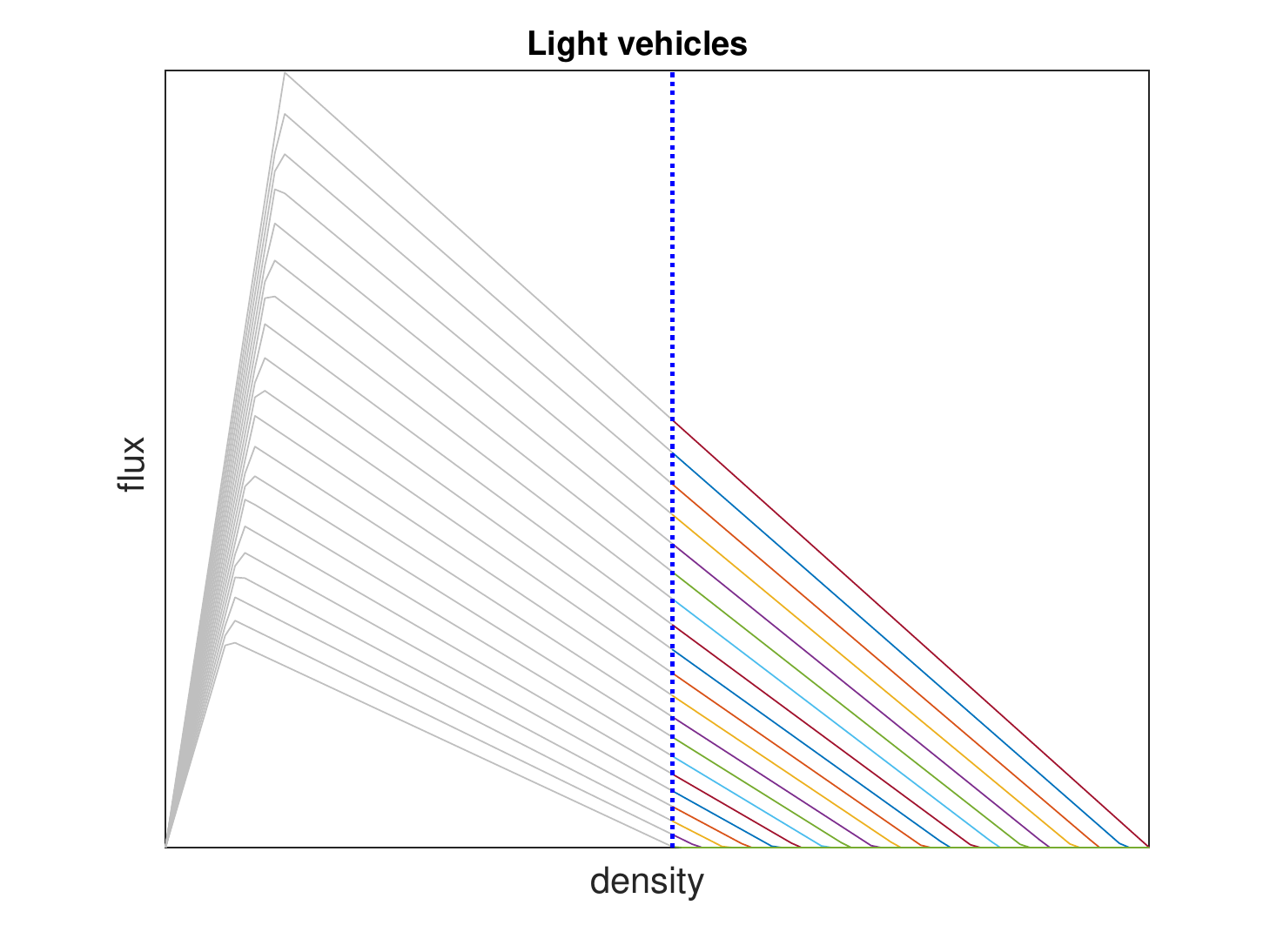}
				\put(50,47){\rotatebox{90}{\tiny transition level}}
			\end{overpic}
			\includegraphics[width=.49\linewidth]{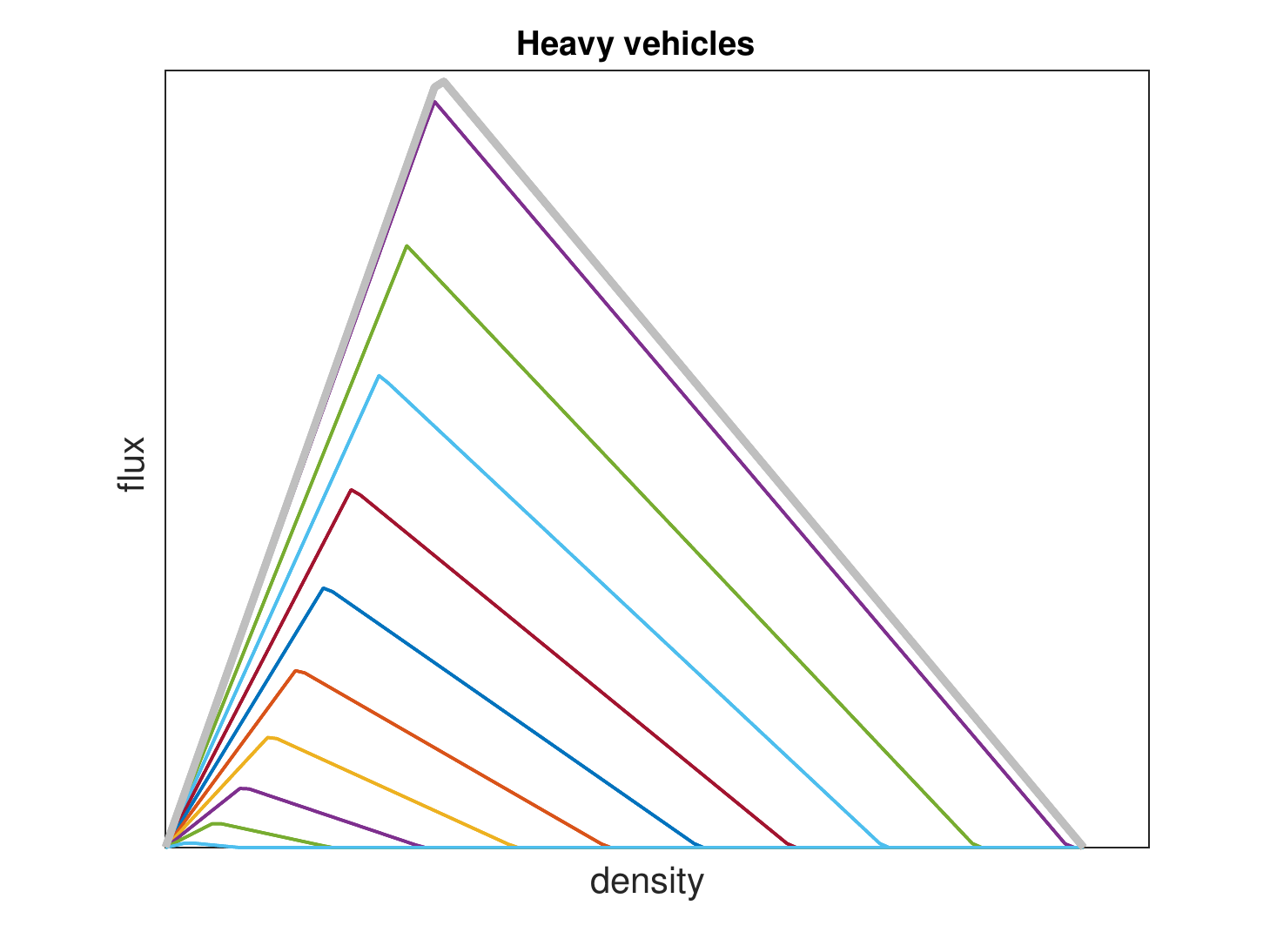}
			 \caption{Families of fundamental diagrams in the two phases. Left: function $\rhoL\to f_\leg(\rhoL,\rhoP)$ for several values of $\rhoP$ in the case of partial (top) and full (bottom) coupling. Right: function $\rhoP\to f_\pes(\rhoL,\rhoP)$ for several values of $\rhoL$ in the case of partial (top) and full (bottom) coupling.}
             \label{fig:DFvariabili}
\end{figure}

Figure~\ref{fig:DFvariabili} shows the families of fundamental diagrams devised for taking into account the flux-density dependence for each class of vehicles given the density of the other class, see \cite{briani2021AXIOMS} for more details.

\medskip

We are now ready to describe the test. Let us consider a stretch of road of length 32 km, across two segments. 
A sensor is located at the interface at $x=20.5$ km.
A permanent bottleneck caused by the transition from 3 to 2 lanes is located at $x=29.5$ km.

In the real scenario, confirmed by direct observation of Autovie Venete personnel, the bottleneck causes congestion for heavy vehicles only, which propagates back and, in turn, slows down the light vehicles.
As in the academic test, the model perceives the congestion only when it reaches the sensor.

\begin{figure}
    \centering
    \includegraphics[width=.4\linewidth]{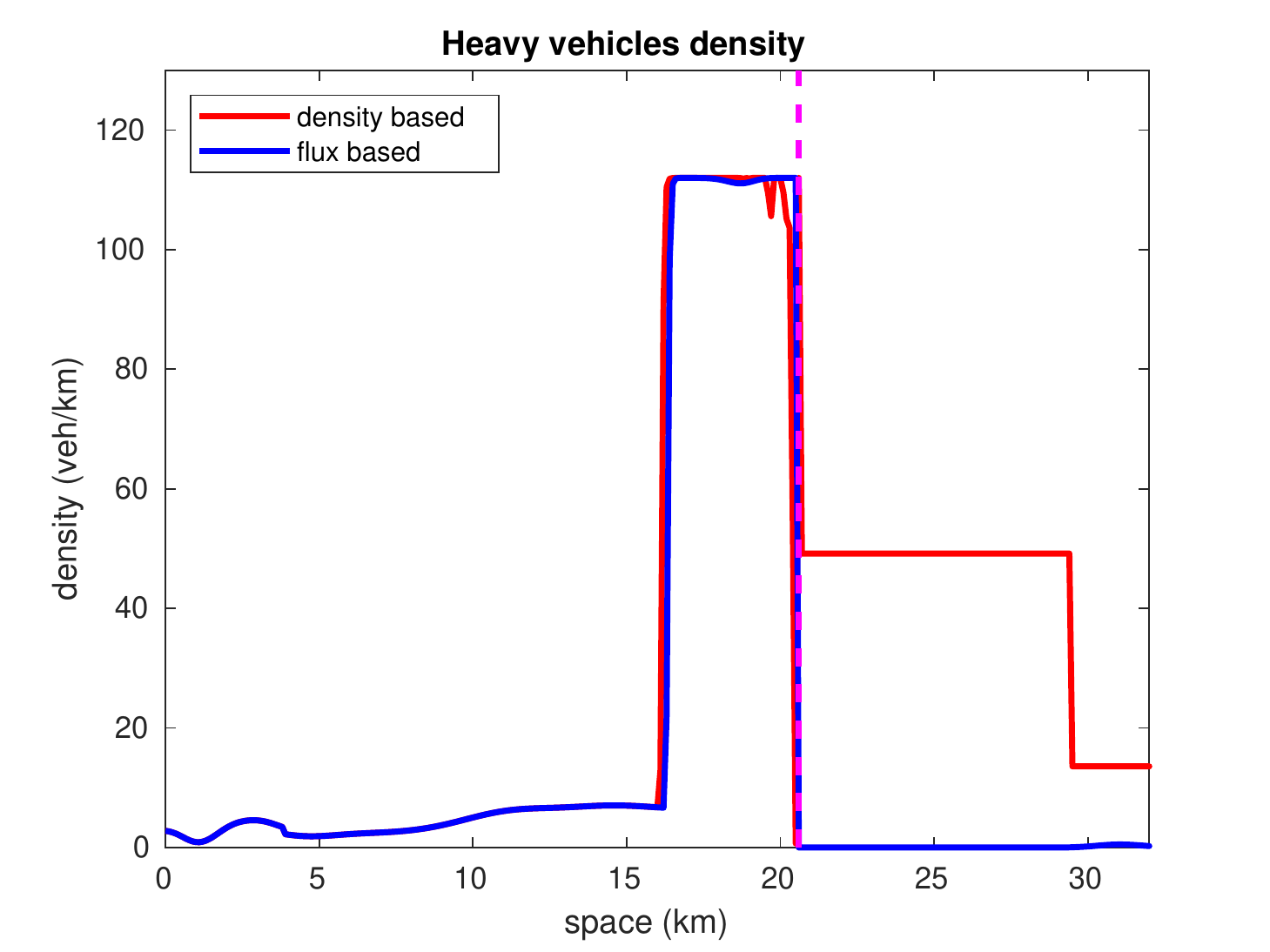}
    \includegraphics[width=.4\linewidth]{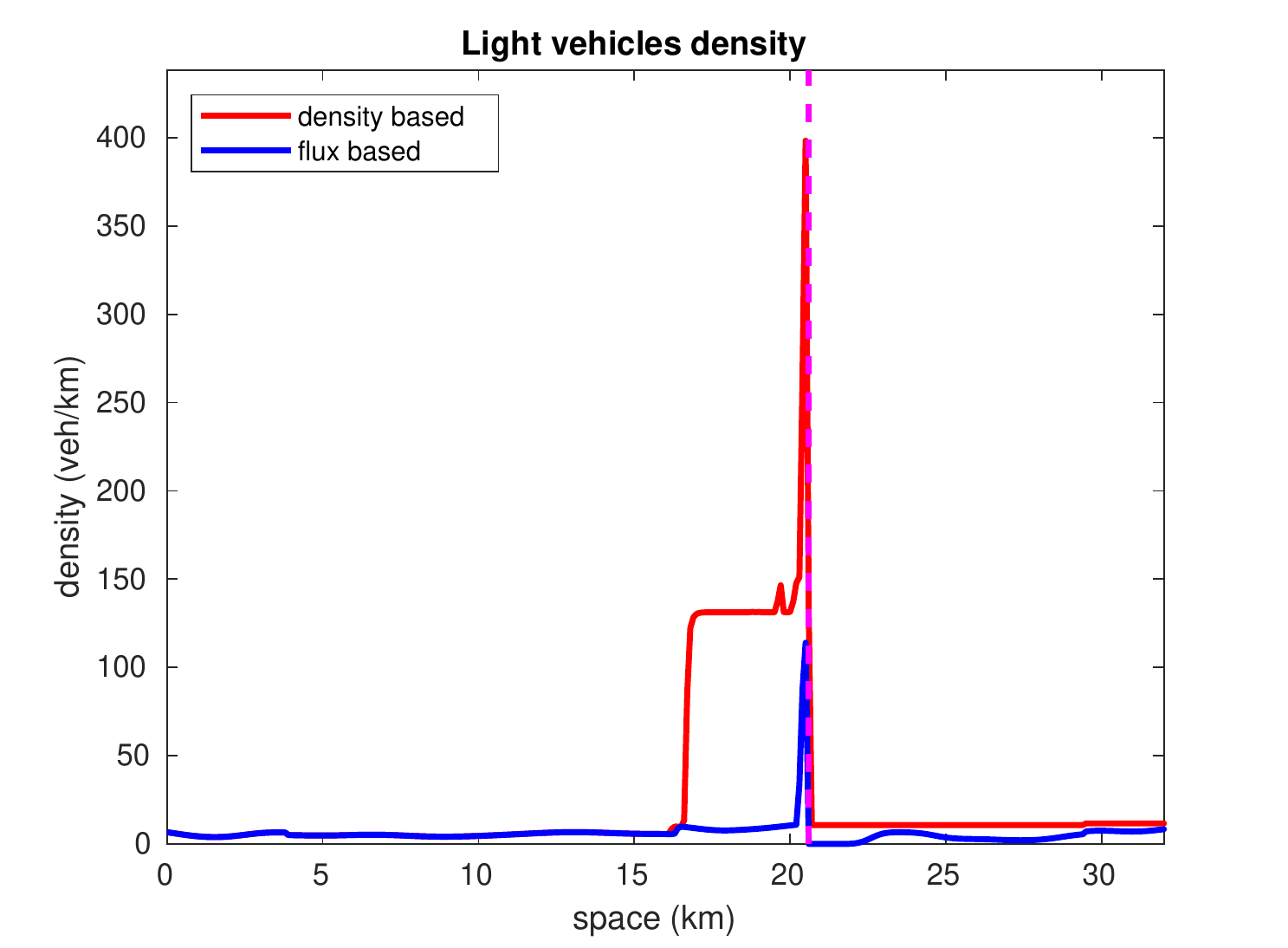}\\
     \includegraphics[width=.4\linewidth]{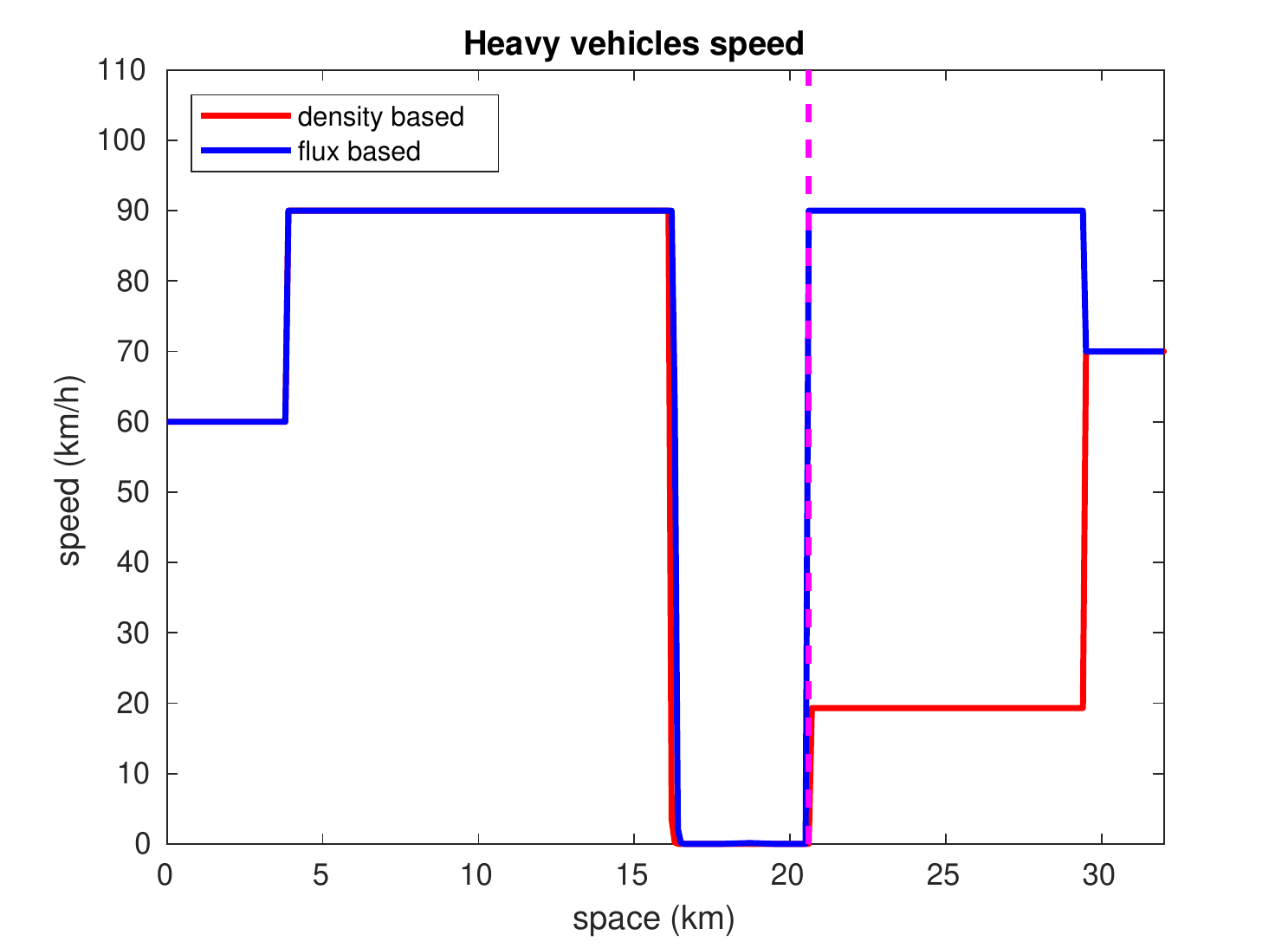}
    \includegraphics[width=.4\linewidth]{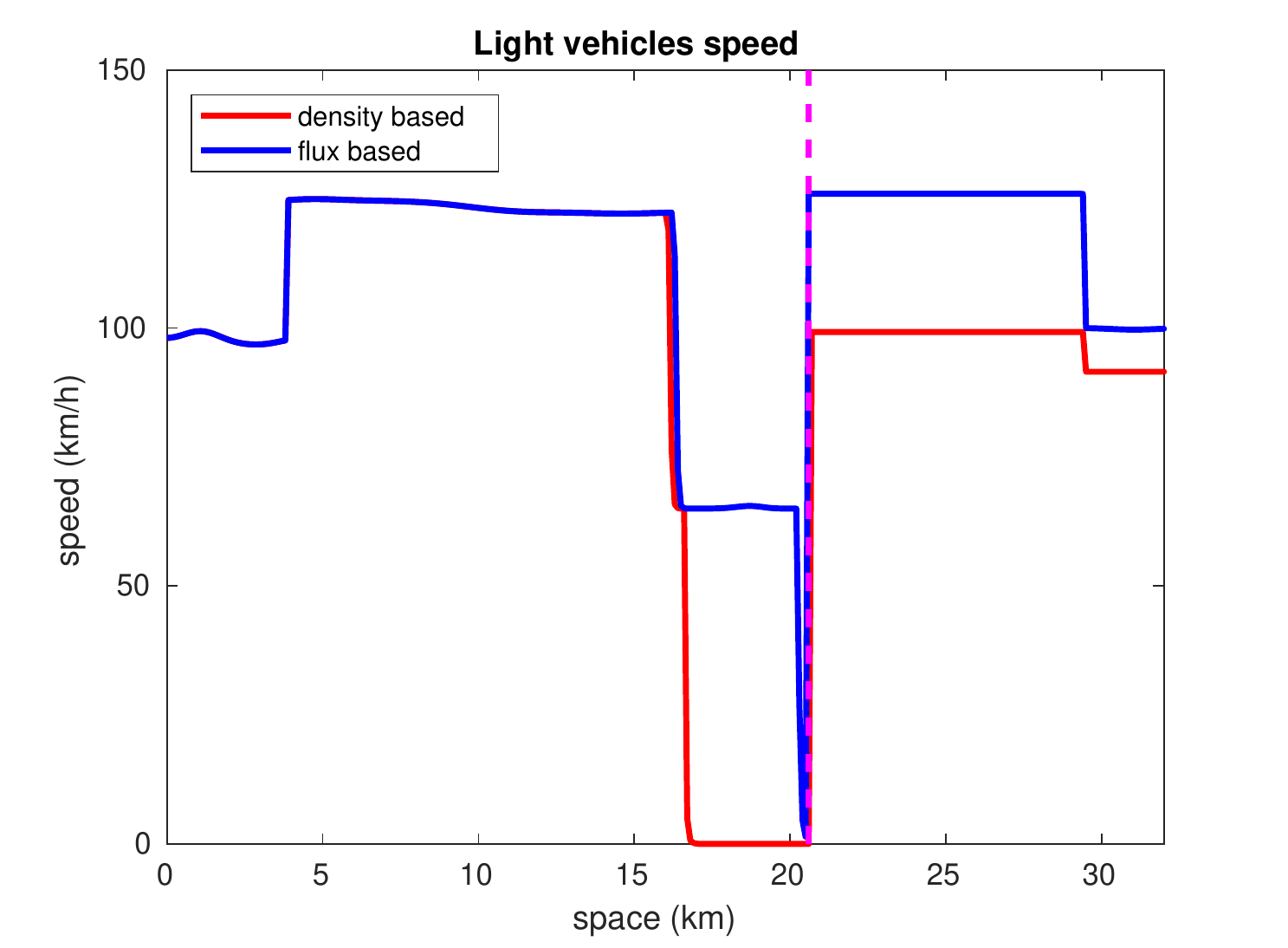}
    \caption{Test with real data. Density and velocity of light and heavy vehicles for the flux- and density-based approach at time $t = t_0$.}
    \label{fig:test_flunas_real_data}
\end{figure}

Figure~\ref{fig:test_flunas_real_data} shows the density and the velocity of light and heavy vehicles in the case of flux- and density-based approaches.
If the flux-based approach is used, a piece of congestion for heavy vehicles propagates upstream the sensor and it keeps slowing down the light vehicles (their speed is about 60 km/h), while downstream traffic is free.
If instead the density-based approach is used, the dynamics are more complicated because the corrections $\rho\to \rho'$ for each class changes the fundamental diagram of the other class, see Figure~\ref{fig:DFvariabili}: downstream, we observe a slowdown for both vehicle classes (velocity is about 20 km/h for heavy vehicles and 100 km/h for light vehicles); upstream, both vehicle classes are totally congested (velocity is zero).
Again, the density-based approach better matches the real scenario especially between the sensor and the bottleneck.

\subsection{Forecast}
In this section, we explore the possible advantage of estimating the incoming traffic volume for traffic forecast. 
The idea is to run again the simulator based on the model \eqref{eq:modelLH} from time $t_0$ to time $t_0+\Delta \tfut$. 
Conversely to the previous case, here we aim at forecasting the traffic distribution in the whole road $[\xmin,\xmax]$ (with no interruption at sensors), starting from the outcome of the nowcast procedure as initial condition for the densities $(\rhoL,\rhoP)$.
Obviously, sensor data cannot be longer used here since they are not yet available, and the problem arises which boundary conditions should be used. 
Ideally, one should estimate the correct future inflow and outflow at each time step $\Delta t$, but this is extremely difficult considering the high variability of the traffic dynamics. 
On the opposite side, the simplest solution is to assume $\rhoin=0$ so as to assume that nobody enters the road, and $\rhoout=0$ so as to guarantee maximal outflow, but this leads to a gradual emptying of the road starting from the inflow boundary. 
A possible compromise is to set $\rhoout=0$ for maximal outflow and keep a \emph{constant} inflow, equal to the last available datum, or an average of the last minutes, or equal to a certain value predicted by a separate procedure. 
Here we consider the outcome of the ANN set up in Section \ref{sec:preconbo}, which estimates the traffic volume for 30 min in the future, directly injecting the flux datum in the numerical scheme.

In order to measure the error of the simulation, we compare the forecast traffic density distribution $x\to \rho^F_{\{L,H\}}(x,t_0+\delta)$ at any future time $t_0+\delta$, $\delta\in[0,\Delta \tfut]$, with the estimated nowcast $x\to \rho^N_{\{L,H\}}(x,t_0+\delta)$ computed as soon as the data becomes available (i.e.\ at time $t_0+\delta$). 
Relative $L^1$-distance between traffic densities is computed as usual
$$
E_{\{L,H\}}(t):=\frac{  \|  \rho_{\{L,H\}}^F  - \rho_{\{L,H\}}^N    \|_{L^1}}{\|\rho_{\{L,H\}}^N\|_{L^1}}
\qquad \text{where} \qquad 
\|\rho\|_{L^1}:=\int_{\xmin}^{\xmax}\left|\rho(x,t)\right|dx.
$$

\begin{figure}
        \centering
        \includegraphics[width=.49\linewidth]{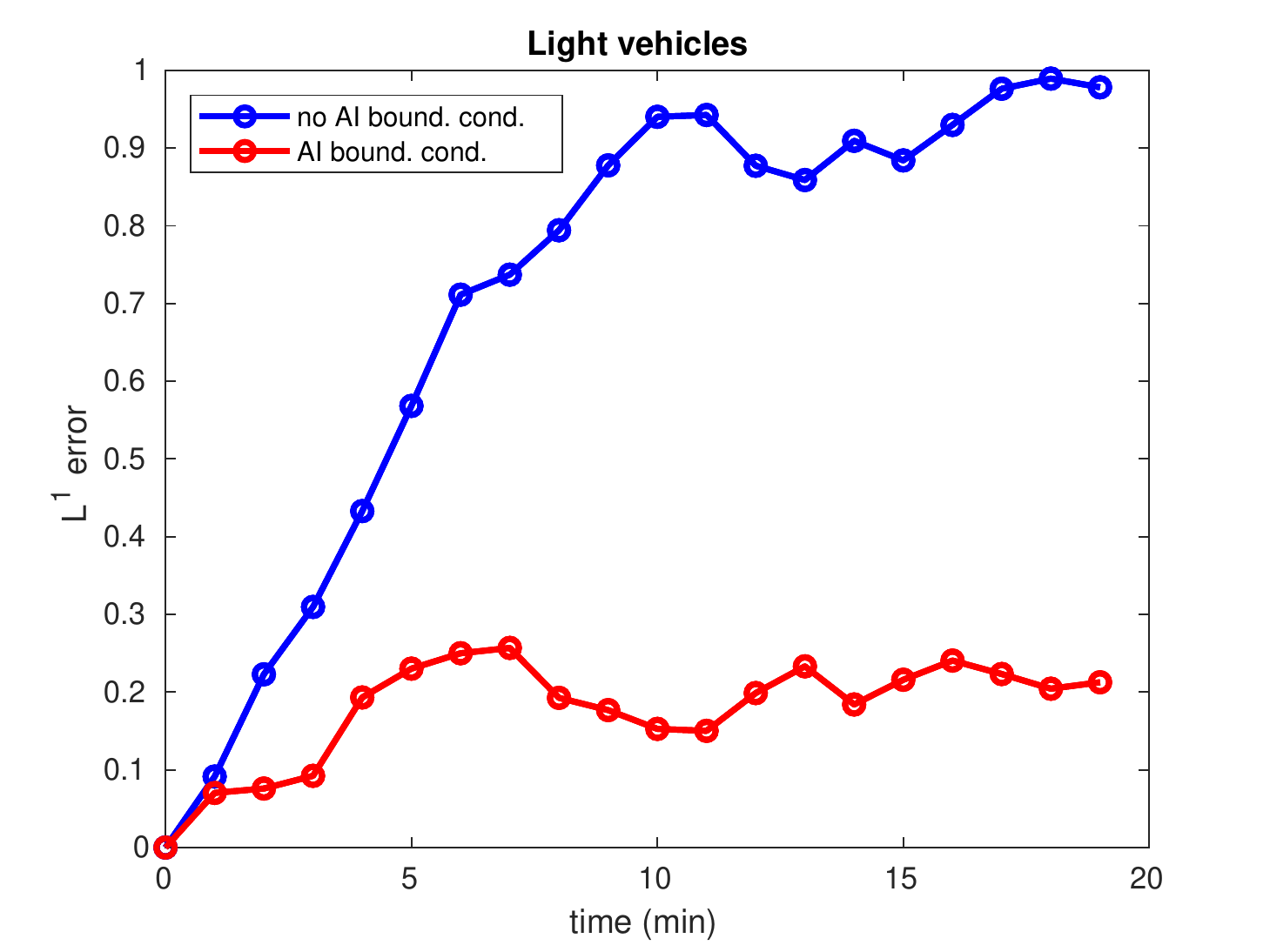}
        \includegraphics[width=.49\linewidth]{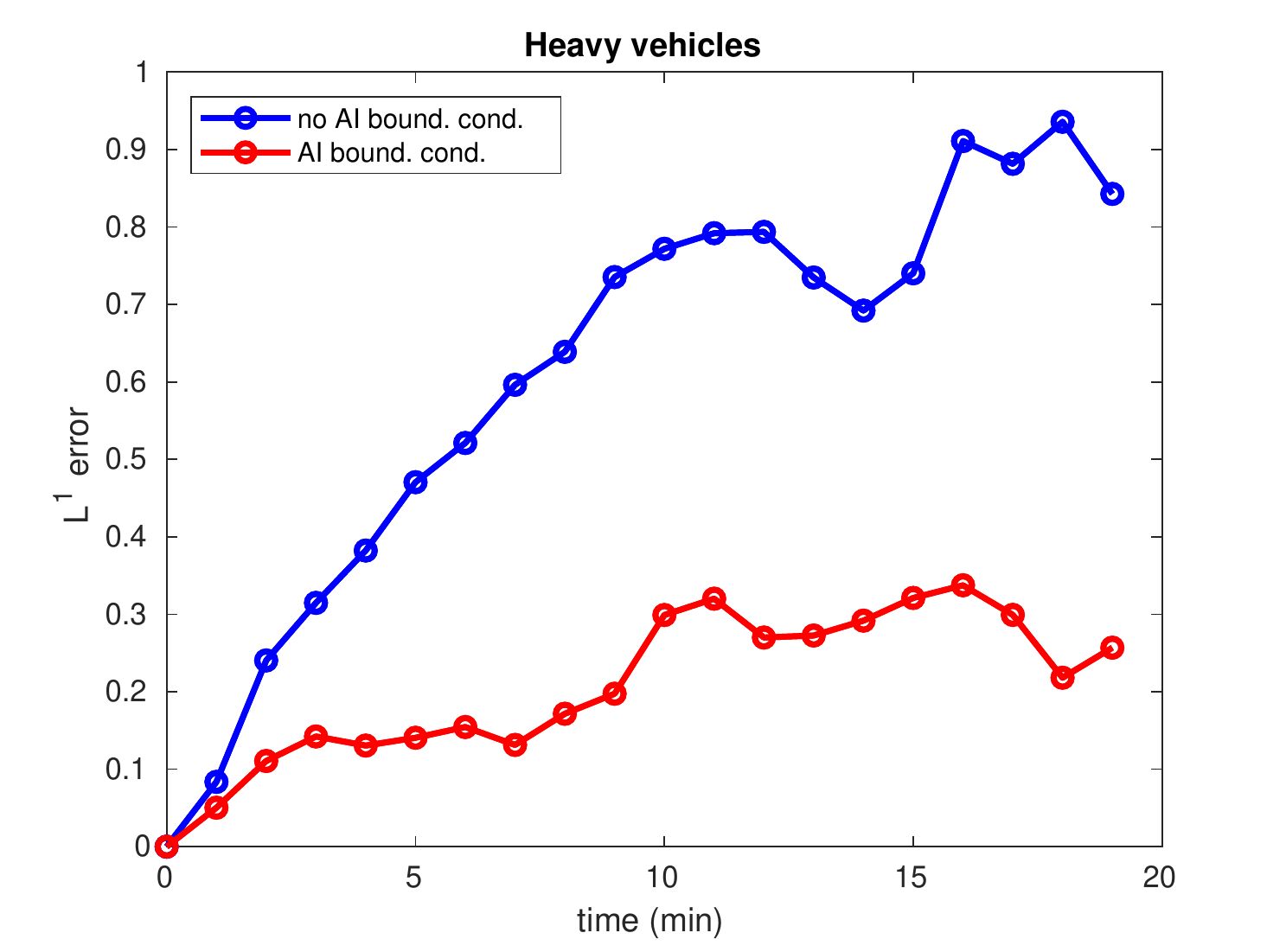}
        \caption{Normalized $L^1$ error as a function of time with ANN-generated inflow boundary condition (red line) and with null inflow (blue line), for light (left) and heavy (right) vehicles.} 
        \label{fig:test_preconbo}
    \end{figure}
Figure~\ref{fig:test_preconbo} shows the error as a function of time for light and heavy vehicles separately, on a stretch of road of length 16 km with two lanes. We compare the error made by using the predicted value of the incoming flux with the one made by simply assuming a null inflow (the simplest choice). 
It is interesting to note that in the ANN-aided traffic prediction, the error initially increases and then stays bounded below about 30\% for both light and heavy vehicles. 
This means that, although the traffic distribution can be shifted horizontally with respect to the real one, the total mass (i.e.\ the amount of vehicles) does not differ excessively. 
On the contrary, using a null inflow the error rapidly increases up to 100\% and then stabilizes, as expected.


\section{Conclusions}\label{sec:conclusions}
In this paper, we have proposed a hybrid model-/data-driven method for reconstructing and predicting traffic distributions on extra-urban roads and highways. Similar to \cite{shi2021IEEE-ITS, herty2022pp}, the idea is to use an ANN as a joining link between real data and the mathematical model, avoiding using the latter in the training phase of the ANN. 
This approach allows exploiting the power of ML in extrapolating information from real-time and historical data and then passing to the model a piece of processed information that can be immediately incorporated.
We have also observed that data-driven approaches based on data measured by fixed sensors can hardly extrapolate any kind of spatial information, i.e.\ information about the spatial distribution of traffic \emph{between} sensors. This is the reason why we think that the mathematical model is essential in TSE since it catches the right causality of traffic dynamics in space and time.

\section*{Acknowledgments} 
The authors want to warmly thank Andrea Appella and Giulia Tatafiore for their support in analyzing and understanding data, as well as improving the quality of the labeled dataset.
The authors want also to thank Paolo Ranut who supported this research on behalf of the motorway company Autovie Venete S.p.A.

\section*{Funding}
This work was carried out within the research project ``SMARTOUR: Intelligent Platform for Tourism'' (No.\ SCN\_00166) funded by the Ministry of University and Research with the Regional Development Fund of European Union (PON Research and Competitiveness 2007--2013). 

E.C.\ and M.B.\ would also like to thank the Italian Ministry of Instruction, University and Research (MIUR) to support this research with funds coming from PRIN Project 2017 (No.\ 2017KKJP4X entitled ``Innovative numerical methods for evolutionary partial differential equations and applications'').

This work was also partially funded by Autovie Venete S.p.A.

All the authors are members of the INdAM Research group GNCS.

\bibliographystyle{abbrv}
\bibliography{biblio}

\end{document}